\documentclass[sigconf,nonacm]{acmart}
\AtBeginDocument{%
  }

\usepackage{multirow}
\usepackage{enumitem}
\usepackage{url}

\begin{document}

\title{Task-Aware Harness Provisioning for LLM Agents in Mission-Critical Infrastructure Operations}

\author{Liangtao Lin}
\affiliation{%
  \institution{Nanyang Technological University}
  \country{Singapore}
}
\email{liangtao002@e.ntu.edu.sg}

\author{Qingang Zhang}
\affiliation{%
  \institution{Nanyang Technological University}
  \country{Singapore}
}
\email{qingang.zhang@ntu.edu.sg}

\author{Zhaomeng Zhu}
\affiliation{%
  \institution{Nanyang Technological University}
  \country{Singapore}
}
\email{zhaomeng.zhu@ntu.edu.sg}

\author{Tianwei Zhang}
\affiliation{%
  \institution{Nanyang Technological University}
  \country{Singapore}
}
\email{tianwei.zhang@ntu.edu.sg}

\author{Yonggang Wen}
\affiliation{%
  \institution{Nanyang Technological University}
  \country{Singapore}
}
\email{ygwen@ntu.edu.sg}

\renewcommand{\shortauthors}{}


\begin{abstract}
LLM agents have been widely adopted to operate mission-critical infrastructure (MCI). These agents normally rely on a \emph{harness} that determines what information they can access, which tools they can use, and what actions they can take. Existing systems often expose the same comprehensive harness to every task, which may not be necessary and cause resource wastes. In this paper, we focus on the identification of optimal harness configurations, and view it as a resource-matching problem between what each task requires and what the harness provides. To measure this match, we classify MCI tasks based on the mathematical representation of the underlying system and rank harness configurations by the amount and type of information they provide. We then construct task-to-harness mappings from two sources: mining research literature and measuring controlled agent execution.   
Leveraging the measured mapping, we propose a new harness provisioning algorithm: map-guided escalation. It begins with a task-specific harness and expands to full provision only after a failed self-check. We evaluate our method in two representative MCI tasks: in liquid cooling, it improves the agent accuracy from $0.652$ under full provision to $0.715$ and achieves accuracy comparable to Reflexion with $48\%$ fewer tokens; In power grids, full provision remains accuracy-optimal, while map-based provisioning offers lower-cost alternatives. These findings show that harness provisioning follows a domain-dependent accuracy--cost Pareto frontier rather than a universal optimum.
\end{abstract}

\ccsdesc[500]{Computing methodologies~Intelligent agents}
\ccsdesc[500]{Applied computing~Physical sciences and engineering}
\ccsdesc[500]{Computing methodologies~Simulation evaluation}

\settopmatter{printacmref=false}
\setcopyright{none}
\renewcommand\footnotetextcopyrightpermission[1]{}


\maketitle

\section{Introduction}
Mission-critical infrastructure (MCI), including information technology, power, and water systems, comprises systems and assets whose disruption or destruction could severely affect security, the economy, public health, or public safety~\citep{cisa_critical_infrastructure}. Operating and maintaining (O\&M) these systems require continuous monitoring, fault diagnosis, future-state prediction, maintenance planning, and control. Recent studies have explored LLM agents for cloud operations and industrial asset management~\citep{chen2025aiopslab,patel2025assetopsbench}. To perform these tasks, an agent relies on a \emph{harness} that provides task-relevant information, tools, and system access~\cite{li2026agentharness, meng2026agentsurvey}. It is thus critical to configure the optimal harness for each given task. 

We frame this as a resource-matching problem: a harness determines the information and capabilities available to the agent, while each task imposes its own requirements for reliable execution. The simplest one-size-fits-all harness strategy, which provides every task with the same configurations, has been questioned~\citep{yu2026skill}. More advanced solutions are proposed, e.g., adapting the harness by retrieving resources relevant to the current task~\citep{qin2024toolllm,mudunuri2026semantic}, granting access according to predefined rules~\citep{shi2025progent,kim2025prompt}, or allowing the agent to decide which resources it needs~\citep{jiang-etal-2023-active,asai2024self}. However, these approaches usually infer what may be useful rather than measure what is actually sufficient. In MCI O\&M, provisioning must balance not only task success but also economic, including token usage, latency, and computation, as well as information security by limiting unnecessary data exposure. Too little provision may prevent reliable execution, whereas excessive provision increases cost and exposure. This motivates our central question: \textit{can we determine the minimum harness a given task needs for reliable execution}?

We hypothesize that each task category has a characteristic harness demand, and seek to identify the mapping between them, which we call the \emph{task-to-harness mapping}. To this end, we first need two comparable representations: one for task demand and one for harness provision. MCI O\&M tasks are commonly described using labels such as detection, diagnosis, and prediction, but these labels lack consistent definitions across domains and studies, and do not systematically cover the full task space. We therefore derive a new task taxonomy from the mathematical representation of the underlying physical system, yielding categories with explicit and distinguishable boundaries. On the provision side, we separately define a cumulative harness hierarchy according to the amount and type of system information made available to the agent. Together, these representations provide two structured spaces for task types and resource provision. 

We propose two complementary strategies to identify the task-to-harness mapping. First, we examine if it is already implicit in prior MCI O\&M research. We collect and analyze more than 1{,}200 papers published over the past decade and find a clear pattern: tasks concerning future outcomes, latent states, or interventions tend to use more extensive harness resources. Second, to verify whether this literature-derived pattern reflects actual agent requirements, we conduct controlled execution experiments in two MCI cases: liquid cooling and power grids. We build simulation-backed agent environments with multiple harness levels and a benchmark of 240 verified tasks. Execution results partly align with the literature-derived pattern but also reveal domain-specific variation. In liquid cooling, several task categories achieve best performance below full provision while using fewer tokens and shorter time, showing that the most comprehensive harness is not always optimal.

With the task-to-harness mapping, an MCI agent can provision resources according to the type and estimated demand of each atomic task. The simplest solution is to directly retrieve the corresponding harness level from either the literature-derived or execution-derived map, requiring no additional routing call. To account for variation among tasks within the same category, we further propose a \emph{map-guided escalation} strategy: the agent starts with the execution-derived provision and retries with the full harness only when the initial execution fails its self-check. On held-out tasks, execution-map lookup performs comparably to experience-augmented LLM routing while using $14\%$ and $12\%$ fewer tokens than full provision. Map-guided escalation improves accuracy over full provision from $0.652$ to $0.715$ in liquid cooling. It also achieves comparable accuracy to Reflexion, an iterative method that uses verbal self-reflection on task feedback to improve subsequent attempts~\citep{shinn2023reflexion}, while using $48\%$ fewer tokens. In the power-grid domain, full provision remains accuracy-optimal, while execution-map lookup provides a lower-cost operating point. Overall, these results reveal a domain-dependent Pareto frontier between execution accuracy and resource cost rather than a single provisioning policy that is optimal across systems.

Our contributions are summarized as follows:
\begin{itemize}[noitemsep,leftmargin=*]
    \item We formulate harness provisioning for MCI agents as a resource-matching problem, balancing task performance, execution cost, and unnecessary information exposure.

    \item We introduce a novel MCI agent task taxonomy derived from system equations and an ordered harness hierarchy with increasing system access, providing comparable representations of task demand and harness provision.

    \item We construct two task-to-harness maps through data-driven analysis of more than 1{,}200 MCI papers and controlled execution on a new benchmark of 240 verified tasks in two physical cases.

    \item We propose map-guided escalation, which uses the measured mapping to initialize harness provision and expands access only when needed, improving the accuracy--cost trade-off over fixed and dynamically routed provisioning.
\end{itemize}

\begin{figure*}[t]
  \centering
  \includegraphics[width=0.85\textwidth]{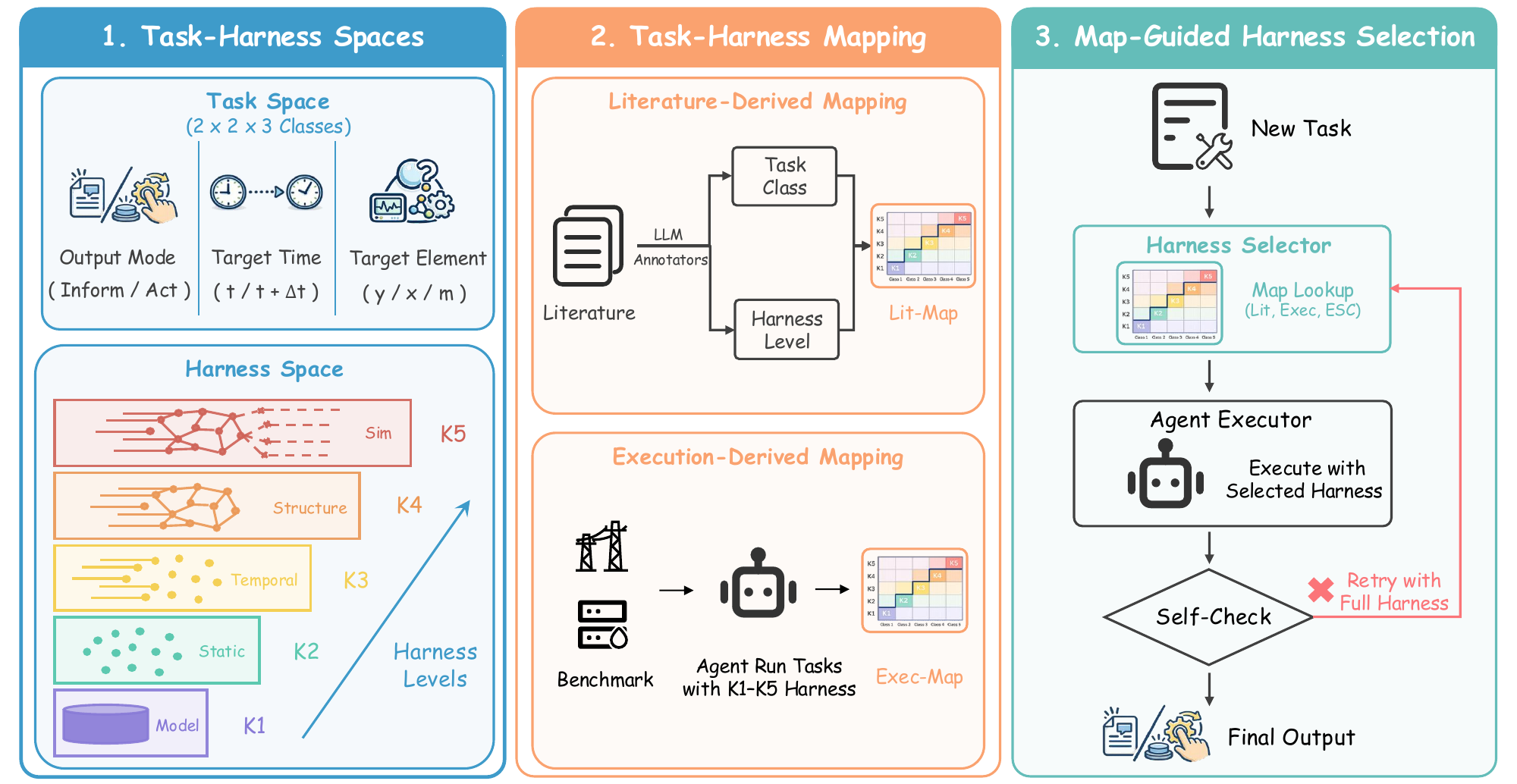} 
  \caption{Overview of the Proposed Framework, which defines the task and harness spaces, estimates literature- and execution-derived task-to-harness maps, and uses the resulting map for harness selection with optional self-check-triggered escalation.}
  \label{fig:method-overview}
\end{figure*}

\section{Related Work}
\subsection{LLM Agents and Benchmarks for MCI}

Operational intelligence in mission-critical infrastructure has traditionally relied on task-specific pipelines. Representative KDD systems include EGADS for scalable anomaly detection~\citep{laptev2015egads}, SR-CNN and OmniAnomaly for time-series monitoring~\citep{ren2019timeseries,su2019omnianomaly}, and CIRCA for causal root-cause analysis~\citep{li2022circa}. These methods provide strong solutions for individual detection or diagnosis tasks, but do not support a general agent across the O\&M lifecycle.

Recent work moves toward interactive agents and executable benchmarks. AIOpsLab deploys fault-injected cloud environments and evaluates agents across the incident lifecycle~\citep{chen2025aiopslab}, while AssetOpsBench provides tools and scenarios for industrial asset operation and maintenance~\citep{patel2025assetopsbench}. In related infrastructure control, LLMLight evaluates LLM agents for traffic-signal decision making~\citep{lai2025llmlight}. Existing surveys emphasize that realistic agent evaluation must cover behavior, reliability, safety, environments, and tooling~\citep{mohammadi2025evaluation}. These benchmarks evaluate agents within predefined environments and interfaces. Our benchmark instead varies harness provision along an ordered ladder, holds the executor fixed, and estimates the sufficient provision for each task class.

\subsection{Improving Agent Performance}

Most prior work improves the execution side of an agent by changing how it reasons, learns, or reacts within a given environment. ReAct interleaves reasoning with actions~\citep{yao2023react}, Reflexion introduces feedback-driven retries~\citep{shinn2023reflexion}, and ExpeL retrieves experience from prior trajectories~\citep{zhao2024expel}. Training-based methods further strengthen planning and tool use: AgentGen generates environments and tasks for planning-oriented instruction tuning~\citep{hu2025agentgen}, while Tool-MVR learns tool invocation and error correction from verified trajectories~\citep{ma2025toolmvr}. These methods primarily improve the executor or its execution process under an available harness.

A complementary line of work improves the supply side by adapting what is exposed to the executor. ToolLLM and AnyTool retrieve task-relevant APIs from large tool collections~\citep{qin2024toolllm,du2024anytool}, while Instruction-Tool Retrieval (ITR) dynamically retrieves relevant instruction fragments and exposes a reduced tool subset at each step \cite{franko2025dynamic}. Chameleon plans compositions of external modules~\citep{lu2023chameleon}, and Sufficient Context predicts whether retrieved textual evidence is adequate for answering a query~\citep{joren2025sufficient}. Model-routing methods such as RouteLLM and AutoMix similarly allocate computational capacity according to predicted need, although they switch executors rather than vary the harness of a fixed executor~\citep{ong2025routellm,aggarwal2024automix}. These approaches select resources through relevance, learned routing, or model-reported need. Our work instead measures the lowest harness provision that preserves execution performance.

\subsection{Harness Design, Evaluation, and Governance}

Recent work increasingly treats the agent harness as a first-class system layer spanning execution control, tool access, context, state, verification, and recovery~\citep{li2026agentharness, meng2026agentsurvey}. Natural-Language Agent Harnesses externalizes control logic into portable specifications~\citep{pan2026naturalharness}. Meta-Harness searches over harness implementations~\citep{lee2026metaharness}. Harness-Bench measures configuration-level harness effects across models and workflows~\citep{yao2026harnessbench}. These studies establish that agent performance depends on the model--harness configuration, but focus on harness representation, optimization, or cross-configuration evaluation.

A parallel body of work governs how harness resources may be used. Progent enforces programmable tool policies~\citep{shi2025progent}, Prompt Flow Integrity constrains information flows~\citep{kim2025prompt}, and AgentSandbox and MiniScope provide sandboxing and permission analysis~\citep{zhang2025agentsandbox,zhu2025miniscope}. ToolPrivBench studies unnecessary privilege selection~\citep{yang2026toolprivbench}, while HarnessAudit evaluates boundary compliance over complete execution trajectories~\citep{liu2026harnessaudit}.Collectively, these studies treat the harness as an object to design, optimize, evaluate, or govern. Our work addresses a complementary question: for a fixed executor and recurring task class, what is the sufficient harness provision that preserves execution performance?

\section{Overview}
\label{sec:overview}

Figure~\ref{fig:method-overview} presents the overall workflow of our methodology, which consists of three stages: defining the task and harness spaces, estimating the relationship between them, and using the resulting mapping to select harnesses for new tasks.

First, to compare what different tasks require with what different harnesses provide, we formally define the task space from the underlying physical system, and organize harnesses into levels with increasing information and capabilities (Section~\ref{sec:framework}).

Second, we establish the task-to-harness mapping from two perspectives (Section~\ref{sec:map-estimation}). We first mine existing MCI O\&M studies to examine which harness levels prior work uses for different task classes. We then construct a controlled benchmark in two MCI environments and execute the same agent under different harness levels to empirically estimate the mapping.

Third, we investigate how the resulting map can be used to select harnesses for new tasks (Section~\ref{sec:selection}). Given a MCI agent task, we first assign it to a task class and retrieve the corresponding harness level from the map. Direct lookup uses this level as the final provision, whereas map-guided escalation uses it as the initial provision and retries with the full harness if the first execution fails its self-check. This allows the class-level map to guide efficient provisioning while accounting for variation among individual tasks.

\section{Task--Harness Spaces}
\label{sec:framework}

Determining which harness a task requires is fundamentally a measurement problem. Before this relationship can be measured, task demand and harness provision must be expressed using stable and comparable representations. We therefore construct two rulers. The first organizes MCI tasks according to what they query or affect in the underlying physical system. The second organizes harness configurations according to the information, tools, and operational access provided to the agent. Together, these rulers turn harness provisioning into a measurable mapping from task class to minimum sufficient harness.

\subsection{Task Space}

Conventional task labels such as detection, diagnosis, prediction, and planning are unsuitable as primary task categories because their meanings vary across domains, applications, and implementations. We therefore derive the task space directly from a common representation of partially observed physical systems.

Let $y_t$ denote observable signals, $x_t$ denote latent physical states, and $m_t$ denote the system structure and governing
mechanisms, including topology, dynamics, constraints, and control logic. We abstract the system as
$$
x_{t+1}=F_{m_t}(x_t,a_t,d_t)+\xi_t,
\qquad
y_t=G_{m_t}(x_t)+\epsilon_t,
$$
where $a_t$ is an intervention, $d_t$ is an external disturbance, and $\xi_t$ and $\epsilon_t$ denote process and observation uncertainty. This representation separates what is directly observed, what remains latent, and what governs system behavior. These three elements provide the possible targets of an MCI task.

Then we represent each task as
$$
T=\langle \omega,\tau,e\rangle,
\quad
\omega\in\{\mathrm{Inform},\mathrm{Act}\},
\tau\in\{t,t+\Delta\},
e\in\{y,x,m\}.
$$
The target element $e$ specifies whether the task concerns observable signals, latent states, or system mechanisms. The target time $\tau$ distinguishes the current system from a future outcome. The output mode $\omega$ specifies whether the agent should report information about the target or produce an intervention intended to affect it. Their Cartesian product yields $2\times2\times3=12$ canonical task classes.

Given the evidence $B_t$ available at task time, the required output takes one of two forms:
$$
o_T=
\begin{cases}
\displaystyle
\hat r_T
=
\rho_T\!\left(
p(e_\tau\mid B_t)
\right),
& \omega=\mathrm{Inform},\\[10pt]
\displaystyle
a^\star
=
\arg\min_{a\in\mathcal{A}}
\mathbb{E}\!\left[
J_T\!\left(e_\tau^{do(a)}\right)
\mid B_t
\right],
& \omega=\mathrm{Act}.
\end{cases}
$$
Here, $p(e_\tau\mid B_t)$ represents the agent's inferred belief about the target, and $\rho_T$ converts this belief into the task-specific report $\hat r_T$, such as a value, label, event, or explanation. For an Act task, $e_\tau^{do(a)}$ denotes the target outcome under intervention $a$, and $J_T$ evaluates the desirability of that outcome. An Inform task therefore reports knowledge about $e$ at time $\tau$, whereas an Act task selects an intervention according to its expected effect on $e$ at that time.

Conventional task types can be represented by specific coordinates in this space according to their target, time horizon, and output mode.
For example, detection corresponds to
$\langle\mathrm{Inform},t,y\rangle$, diagnosis to $\langle\mathrm{Inform},t,m\rangle$, and prediction to $\langle\mathrm{Inform},t+\Delta,y\rangle$. In the following, we use the three-dimensional representation $\langle\omega,\tau,e\rangle$ to denote task types.

\subsection{Harness Space}

A task representation specifies what must be accomplished, but not what information, models, tools, and interfaces are exposed to the executor. Motivated by prior work that organizes system information by temporal
reach and structural scope \cite{endsley1995toward,rasmussen1985role}, we define five cumulative levels of harness provision:
$$
\mathcal{H}_{K1}
\subset
\mathcal{H}_{K2}
\subset
\mathcal{H}_{K3}
\subset
\mathcal{H}_{K4}
\subset
\mathcal{H}_{K5}.
$$

\begin{itemize}[noitemsep,leftmargin=*]
    \item \textbf{$K1$: Model-only reasoning.} The agent receives only the task prompt and relies on its parametric knowledge.

    \item \textbf{$K2$: Static knowledge.} The harness additionally provides fixed resources such as manuals, SOPs, specifications, design documents, and rule bases.

    \item \textbf{$K3$: Temporal observations.} The harness additionally provides historical or real-time telemetry, logs, alarms, and time-series measurements.

    \item \textbf{$K4$: Structure and physics.} The harness additionally provides topology, component relations, governing equations, physical constraints, and control logic.

    \item \textbf{$K5$: Forward simulation.} The harness additionally provides executable simulation, digital-twin rollouts, counterfactual evaluation, and optimization over hypothetical future trajectories.
\end{itemize}

The ordering reflects increasing access to the physical system rather than increasing intrinsic intelligence of the executor. It also does not impose a fixed mapping from task coordinates to harness levels: a future-facing task may be solved from temporal observations, while a current-state task may require structural models or simulation. Although provision increases from $K1$ to $K5$, performance need not improve monotonically because additional context and tools may increase cost, introduce irrelevant evidence, or create unnecessary action opportunities. Thus, $K5$ denotes maximal provision rather than an assumed performance optimum.

\section{Estimating the Task-to-Harness Map}
\label{sec:map-estimation}

The task space characterizes what different MCI tasks demand, while the harness space characterizes what information and capabilities are provided to the agent. Our goal is to identify the relationship between these two spaces by assigning a harness level to each task class. We first define a common mapping rule and then estimate the map from two sources. The literature-derived map summarizes which harness levels prior MCI O\&M studies use for different task classes, whereas the execution-derived map identifies the lowest harness level that achieves near-best performance when the same agent is evaluated under different harness levels.

\subsection{Mapping Definition}
\label{sec:mapping-definition}

Let $A_s(T,K)$ denote the support for assigning harness $\mathcal{H}_K$ to task class $T$ under evidence source $s\in\{\mathrm{lit},\mathrm{exec}\}$. We define
\[
\mathcal{G}_s(T)
=
\min\left\{
K\in\mathcal{K}:
A_s(T,K)
\geq
\max_{K'\in\mathcal{K}}A_s(T,K')-\epsilon_s
\right\},
\]
where $\epsilon_s\geq0$ is a source-specific tolerance. The mapping selects the lowest harness level whose support is within $\epsilon_s$ of the maximum for that task class, avoiding unnecessary provision when several levels receive similar support.

For the literature-derived map, we set $A_{\mathrm{lit}}(T,K)=P_{\mathrm{lit}}(K\mid T)$, where $P_{\mathrm{lit}}(K\mid T)$ is the observed frequency of harness level $K$ among prior studies assigned to task class $T$. The resulting map $\mathcal{G}_{\mathrm{lit}}$ summarizes how prior work has provisioned different task classes.

For the execution-derived map, we set $A_{\mathrm{exec}}(T,K)=\mu_{\mathrm{exec}}(T,K)$, where $\mu_{\mathrm{exec}}(T,K)$ is the mean execution score obtained under harness level $K$ for task class $T$. The resulting map $\mathcal{G}_{\mathrm{exec}}$ selects the lowest harness level whose measured performance is within $\epsilon_{\mathrm{exec}}$ of the best observed performance.

\subsection{Literature-Derived Mapping}
\label{sec:literature-map}

We first examine which harness levels existing MCI O\&M studies use for different task classes. We collect approximately 2,000 candidate papers from \textit{arXiv}, \textit{Semantic Scholar}, and \textit{OpenAlex}, and retain more than 1,200 studies within our scope. For each paper $p$, three LLM annotators independently extract its primary task class $T_p\in\mathcal{T}$ and the highest harness level $K_p\in\mathcal{K}$ materially used by the proposed method. The labels are consolidated by majority vote, with unresolved disagreements manually adjudicated.\footnote{Appendix~\ref{app:literature_analysis} details the collection, screening, and annotation procedures.}

We then estimate
\[
P_{\mathrm{lit}}(K\mid T)
=
\frac{N_{\mathrm{lit}}(T,K)}
{\sum_{K'\in\mathcal{K}}N_{\mathrm{lit}}(T,K')},
\]
where $N_{\mathrm{lit}}(T,K)$ is the paper count associated with task class $T$ and harness level $K$. Applying the common mapping rule produces $\mathcal{G}_{\mathrm{lit}}$.

The above map captures established provisioning practice rather than verified sufficiency. A frequently-used harness level may still be unnecessary, insufficient, or over-provisioned for a particular agent or environment. Nevertheless, it provides a literature-scale view of how harness provision varies across task classes.

\begin{figure*}[t]
  \centering
  \includegraphics[width=.85\textwidth]{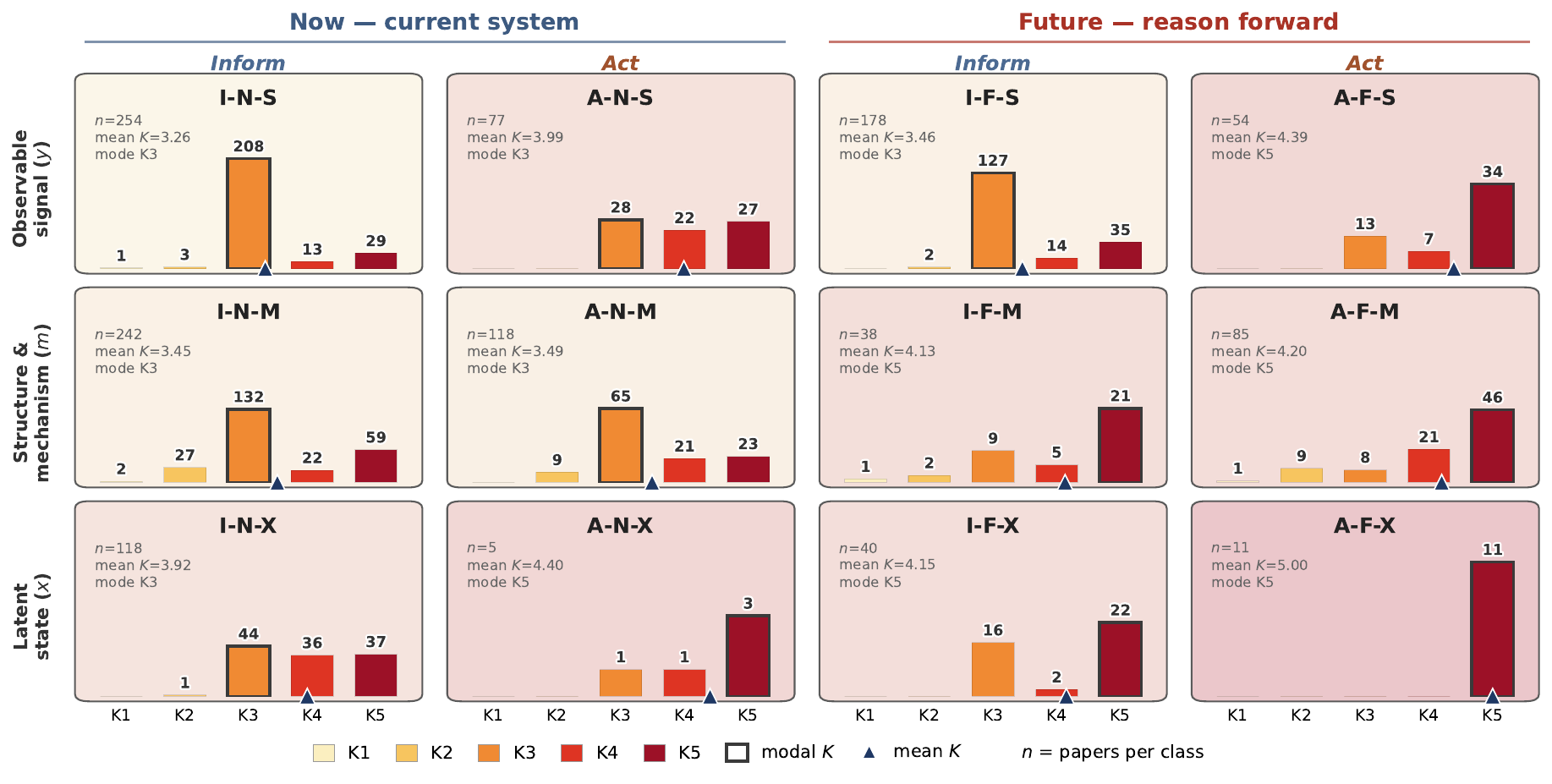}
  \caption{\textbf{Literature-derived Harness Distributions across the 12 Task Classes.} Based on 1,220 MCI O\&M papers, each panel shows the number of papers assigned to harness levels $K_1$--$K_5$ for one task class; the outlined bar marks the modal level while the triangle and background marks the mean. The distributions generally shift toward higher harness levels for Act, Future, and latent-state tasks, while most classes peak at either $K_3$ or $K_5$.}
  \label{fig:lit_dist}
\end{figure*}

\subsection{Execution-Derived Mapping}
\label{sec:execution-map}

We next measure the relationship directly by evaluating the same agent under different harness levels\footnote{Appendix~\ref{app:harness_implementation} details the domain-specific
harness implementation.}. We construct two simulation-backed environments: a thermal--hydraulic digital twin of a liquid-cooled data hall and a modified IEEE-14 power-grid environment built with Grid2Op and pandapower~\citep{marot2021learning,pandapower.2018}. Each environment generates replayable system trajectories from which we instantiate ten tasks for each of the 12 task classes, yielding 240 tasks in total. The tasks cover reporting current observations, inferring latent states or system mechanisms, predicting future outcomes, and selecting interventions. Each task is paired with a deterministic simulator-backed oracle and undergoes both programmatic and human verification.\footnote{Appendix~\ref{app:benchmark} details the benchmark construction and verification.}

Every task is executed under all five harness levels using the same agent, reasoning protocol, and evaluation procedure; only the available harness provision varies. Let $\mu_{\mathrm{exec}}(T,K)$ denote the mean score of harness level $K$ over tasks in class $T$. Applying the common mapping rule gives
\[
\mathcal{G}_{\mathrm{exec}}(T)
=
\min\left\{
K\in\mathcal{K}:
\mu_{\mathrm{exec}}(T,K)
\geq
\max_{K'\in\mathcal{K}}\mu_{\mathrm{exec}}(T,K')
-\epsilon_{\mathrm{exec}}
\right\}.
\]

The above execution-derived map identifies the lowest harness level that achieves near-best performance when the same agent is evaluated across harness levels.

\begin{figure*}[t]
  \centering
  \includegraphics[width=0.85\textwidth]{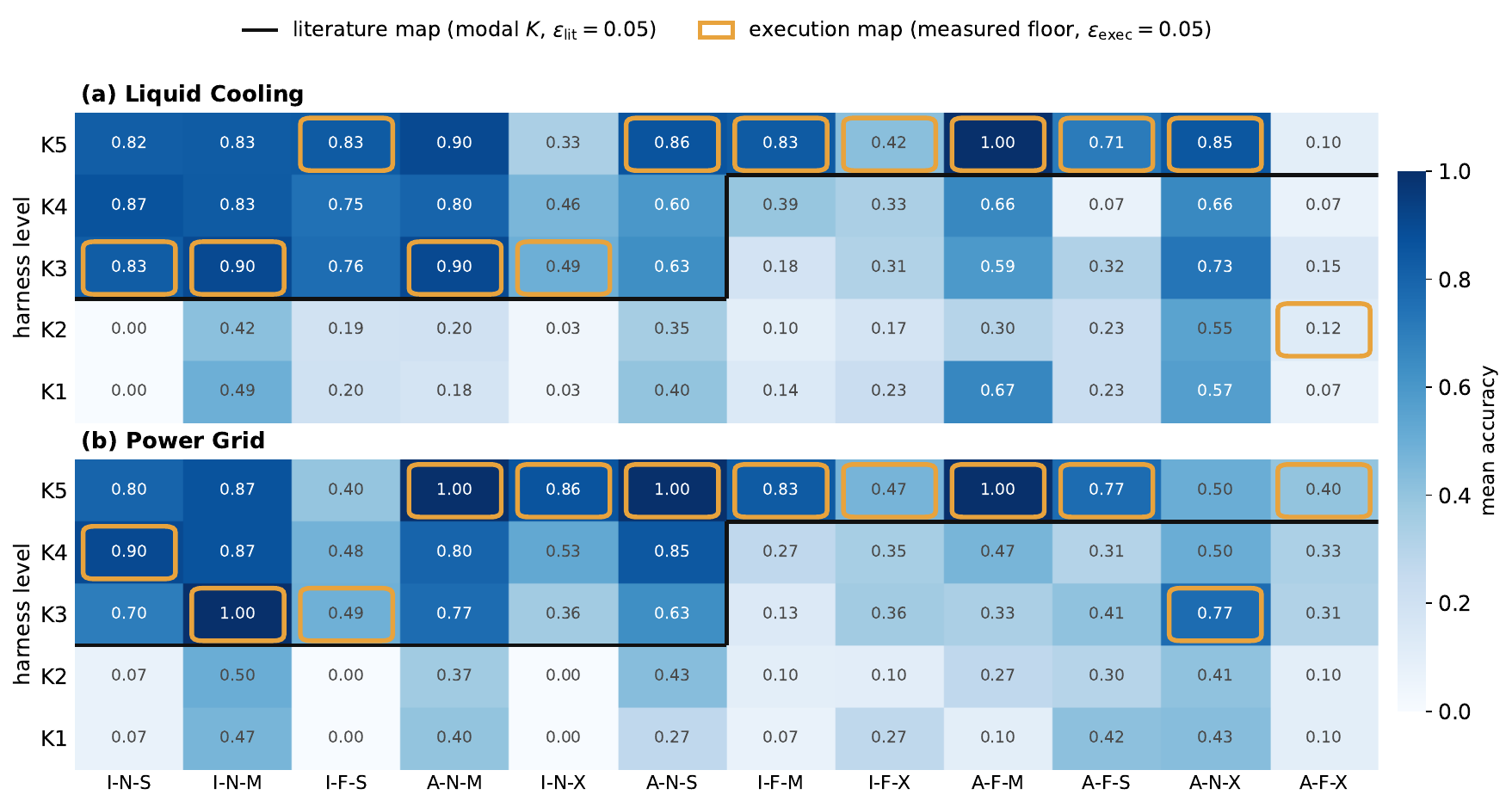} 
  \caption{\textbf{Task-to-Harness maps.} Cells report the mean construction-split accuracy for each task class and harness level (5 tasks per class, 3 runs). Amber boxes mark the lowest level within $\epsilon_{\mathrm{exec}}=0.05$ of the best execution score, while the black staircase marks the literature-derived level. Columns are ordered by the literature mean harness level.}
  \label{fig:task_harness_map}
\end{figure*}

\section{Map-Guided Harness Selection}
\label{sec:selection}

Given an atomic task $q$, the system first assigns it to a task class $T(q)$ and uses the task-to-harness map to select an initial harness. The executor then completes the task using only the information and capabilities exposed by the selected harness. Because tasks within the same class may still differ in difficulty and required information, we allow the harness to expand to $K_5$ when the initial execution fails its self-check.

\noindent\textbf{Initial harness selection.}
Since the task-to-harness map captures recurring provisioning patterns for each task class, we use the mapped level as a class-level prior for selecting the initial harness. Given a task $q$ and a map $\mathcal{G}_s$, where $s\in\{\mathrm{lit},\mathrm{exec}\}$ denotes the map source, the initial harness level is
\[
K_0(q;s)=\mathcal{G}_s(T(q)).
\]
Once the task class $T(q)$ is available, this step requires only a deterministic table lookup and introduces no additional routing call. Let $\pi$ denote the fixed executor. Running $\pi$ under $\mathcal{H}_{K_0(q;s)}$ produces
\[
(o_0,c_0,r_0)
=
\pi\!\left(q,\mathcal{H}_{K_0(q;s)}\right),
\]
where $o_0$ is the initial output, $c_0\in\{\mathrm{pass},\mathrm{fail}\}$ is its self-check result, and $r_0$ contains findings that can be reused in a subsequent attempt.

\noindent\textbf{Conditional escalation.}
The mapped level is a class-level prior and may not provide sufficient information or capabilities for every task instance. When the initial execution fails its self-check, we therefore expand the harness to $K_5$, exposing the executor to the full set of information, tools, and operational permissions. The initial result is accepted if it passes the self-check or if the selected harness is already $K_5$; otherwise, the executor retries once under $K_5$ while retaining the useful findings $r_0$ from the initial attempt. The final harness and output are
\[
\bigl(K_{\mathrm{final}}(q),o(q)\bigr)
=
\begin{cases}
\bigl(K_0(q;s),o_0\bigr),
& c_0=\mathrm{pass}\ \text{or}\ K_0(q;s)=K_5,\\[3pt]
\bigl(K_5,\pi(q,\mathcal{H}_{K_5};r_0)\bigr),
& c_0=\mathrm{fail}\ \text{and}\ K_0(q;s)<K_5.
\end{cases}
\]

\noindent\textbf{Method variants.}
This formulation produces three evaluated variants. \textsc{Lit-Lookup} uses $\mathcal{G}_{\mathrm{lit}}$ and disables escalation, while \textsc{Exec-Lookup} uses $\mathcal{G}_{\mathrm{exec}}$ and likewise accepts the mapped harness as final. \textsc{Map-ESC} uses $\mathcal{G}_{\mathrm{exec}}$ for initial selection and applies the self-check-triggered escalation rule above. The two lookup variants isolate the value of the maps themselves, while \textsc{Map-ESC} evaluates whether failures caused by instance-level variation can be recovered through conditional expansion.

We escalate directly to $K_5$ rather than testing each intermediate level to avoid repeated execution overhead; Appendix~\ref{app:escalation_analysis} shows that intermediate retries rarely terminate before reaching $K_5$ in the evaluated domains. The executor and reasoning protocol remain fixed across both attempts, and the harness enforces the capability boundary at each level. Before escalation, the executor therefore cannot access information, tools, or permissions beyond its initially selected harness.

\section{Experiments}
\label{sec:experiments}

Having defined the task-to-harness maps and their deployment policies, we now evaluate whether the proposed structure is empirically supported and operationally useful. Our experiments address four research questions. 
\begin{itemize}[noitemsep,leftmargin=*]
    \item \textbf{RQ1}: Do task classes exhibit distinct harness requirements? Does richer provision consistently improve performance? 
    
    \item \textbf{RQ2}: How reliably can the task-to-harness map be estimated from literature and execution? How do the two evidence sources compare with each other? 
    
    \item \textbf{RQ3}: How should the map be operationalized at deployment time: through direct lookup or map-guided escalation?
    
    \item \textbf{RQ4}: Whether do the resulting task-to-harness relationships generalize across domains and executor models? What is the impact of the granularity, robustness, and deployment implications of the learned map?

\end{itemize}

\begin{table*}[t]
\centering
\caption{\textbf{Harness Provisioning on the Test Split.}
All policies use the same frozen executor and differ only in how the harness is selected. Latency and tokens are normalized to Full-K5. Bold and underline denote the best and second-best deployable results in each column.}
\label{tab:provisioning}
\small
\setlength{\tabcolsep}{3.5pt}
\begin{tabular}{lll ccc ccc}
\toprule
& & &
\multicolumn{3}{c}{\textsc{Liquid}} &
\multicolumn{3}{c}{\textsc{Grid}} \\
\cmidrule(lr){4-6}\cmidrule(lr){7-9}
Group & Policy & Basis
& Acc.\ $\pm$ std & Lat. & Tok.
& Acc.\ $\pm$ std & Lat. & Tok. \\
\midrule

\emph{Fixed}
& Full-K5
& maximal provision
& $0.652\pm0.001$
& $1.00\times$
& $1.00\times$
& $\mathbf{0.806\pm0.009}$
& $\mathbf{1.00\times}$
& $\underline{1.00\times}$ \\

\midrule
\multirow{3}{*}{\emph{Routing}}
& ITR
& semantic relevance
& $\underline{0.670\pm0.015}$
& $0.90\times$
& $0.99\times$
& $0.780\pm0.010$
& $1.20\times$
& $1.02\times$ \\

& LLM-Route
& LLM judgment
& $0.626\pm0.021$
& $\mathbf{0.74\times}$
& $0.93\times$
& $0.744\pm0.024$
& $1.29\times$
& $1.03\times$ \\

& LLM+Exp
& judgment with experience
& $0.655\pm0.010$
& $\underline{0.75\times}$
& $0.96\times$
& $0.766\pm0.011$
& $1.26\times$
& $1.09\times$ \\

\midrule
\multirow{2}{*}{\emph{Cascades}}
& AutoMix
& trained check from K3
& $0.639\pm0.011$
& $1.08\times$
& $1.94\times$
& $\underline{0.785\pm0.029}$
& $3.03\times$
& $2.61\times$\\

& Blind-ESC
& self-check from K1
& $0.664\pm0.026$
& $0.90\times$
& $1.33\times$
& $0.742\pm0.020$
& $5.33\times$
& $1.75\times$ \\

\midrule
\multirow{3}{*}{\emph{Ours}}
& Lit-Lookup
& literature prior
& $0.658\pm0.017$
& $0.84\times$
& $\underline{0.89\times}$
& $0.664\pm0.018$
& $1.30\times$
& $1.01\times$ \\

& Exec-Lookup
& execution-derived map
& \underline{$0.670\pm0.020$}
& $0.83\times$
& $\mathbf{0.86\times}$
& $0.762\pm0.023$
& $\underline{1.13\times}$
& $\mathbf{0.88\times}$ \\

& Map-ESC
& execution map with self-check
& $\mathbf{0.715\pm0.014}$
& $1.03\times$
& $1.15\times$
& $0.782\pm0.010$
& $1.37\times$
& $1.03\times$ \\

\midrule
\multirow{2}{*}{\emph{Reference}}
& Class Oracle
& expected static ceiling
& \emph{0.694}
& ---
& ---
& \emph{0.809}
& ---
& --- \\

& Task Oracle
& expected static ceiling
& \emph{0.783}
& ---
& ---
& \emph{0.843}
& ---
& --- \\

\bottomrule
\end{tabular}
\end{table*}

\subsection{Experimental Setup}
\label{sec:setup}

\noindent\textbf{Environments and tasks.} We evaluate two simulation-backed MCI domains: a proprietary liquid-cooling digital twin (\textsc{Liquid}) and a power-grid environment built with Grid2Op and pandapower (\textsc{Grid}). Each domain contains 120 tasks, with ten instances for each of the 12 classes $T=\langle\omega,\tau,e\rangle$. Within each class, five tasks are used for map construction and five form a disjoint test split. All 240 tasks undergo programmatic and human verification. We perform the full $K_1$--$K_5$ sweep on all tasks, but construct the execution-derived map only from the construction split and evaluate provisioning policies on the test split.\\

\noindent\textbf{Execution protocol.} We instantiate the five cumulative harness levels defined in Section~\ref{sec:framework}. Unless otherwise stated, all conditions use the same frozen GPT-5.4~\citep{openai2026gpt54} executor with a ReAct loop, at most 20 tool iterations, and a 300-second timeout; only harness provisioning varies. Both maps use a tolerance of $\epsilon_{\mathrm{lit}}=\epsilon_{\mathrm{exec}}=0.05$.\footnote{Appendix~\ref{app:map_construction} reports tolerance sensitivity,
selected-level stability, statistical tests, and oracle definitions.}\\

\noindent\textbf{Compared policies.} We compare maximal provisioning (Full-$K_5$), relevance-based retrieval (ITR~\citep{franko2025dynamic}), LLM routing with and without prior experience (LLM-Route, LLM+Exp), a learned cascade (AutoMix~\citep{aggarwal2024automix}), and progressive escalation from $K_1$ (Blind-ESC). Our policies include direct lookup from the literature and execution maps (Lit-Lookup, Exec-Lookup) and two-stage map-guided escalation from the execution-derived mapped level (Map-ESC). We separately compare Map-ESC with Reflexion~\citep{shinn2023reflexion} and ExpeL~\citep{zhao2024expel}, which adapt execution under a fixed $K_5$ harness.\footnote{Appendix~\ref{app:baseline_details} details the implementation and
adaptation of all compared policies.}\\

\noindent\textbf{Metrics.} Task performance is scored in $[0,1]$ using a Gemini-3.1-Pro~\citep{googledeepmind2026gemini31pro} judge combined with rule-based checks. Each condition is run three times, and results are reported as mean $\pm$ population standard deviation. Token counts include routing calls and all escalation attempts; latency and tokens are normalized to Full-$K_5$. We report two non-deployable references: the \textit{Class Oracle} selects the highest-scoring harness for each task class, while the \textit{Task Oracle} selects the highest-scoring harness separately for each task, both using mean scores across the three runs. \footnote{Appendix~\ref{app:experimental_details} details the executor settings, scoring rules, judge configuration, cost accounting, and repetition protocol.}

\subsection{RQ1: Do Task Classes Require Different Harnesses?}
\label{sec:rq1}

Figure~\ref{fig:lit_dist} shows that harness provision in prior MCI O\&M studies varies systematically across task classes. Among 1,200+ papers, current Inform tasks targeting observable signals or system mechanisms are concentrated at $K_3$, whereas Future, Act, and latent-state tasks shift more strongly toward $K_5$. This indicates two recurring provisioning patterns in prior work: observation-driven provision centred on $K_3$ and simulation-backed provision centred on $K_5$.

Figure~\ref{fig:task_harness_map} provides direct execution evidence by evaluating the same executor under $K_1$--$K_5$. \footnote{Appendix~\ref{app:full_results} details the aggregate and per-class $K_1$--$K_5$ sweeps, split-specific analyses, and absolute execution costs.} The lowest level achieving performance within $\epsilon=0.05$ of the best class score varies from $K_2$ to $K_5$ in \textsc{Liquid} and from $K_3$ to $K_5$ in \textsc{Grid}. Performance is also non-monotonic: five classes in \textsc{Liquid} and four in \textsc{Grid} attain their highest score below $K_5$. Thus, harness requirements vary across both task classes and domains, and maximal provision is neither uniformly necessary nor uniformly optimal.

\subsection{RQ2: How Should the Task-to-Harness Map Be Estimated?}
\label{sec:rq2}

Having established that harness requirements vary across task classes, we next compare the two sources used to estimate the map. As shown in Table~\ref{tab:provisioning}, Lit-Lookup selects the level most frequently observed in prior studies, whereas Exec-Lookup selects the lowest level whose construction-split performance is within $\epsilon_{\mathrm{exec}}=0.05$ of the best level for each task class. The two maps perform similarly on \textsc{Liquid} ($0.658$ versus $0.670$), but differ substantially on \textsc{Grid}, where Exec-Lookup achieves $0.762$ compared with $0.664$. Literature evidence therefore provides a useful prior, while execution evidence better adapts the map to the evaluated domain and executor.

Exec-Lookup also provides a stronger accuracy--efficiency balance than per-task routing. Directly asking an LLM to select the harness (LLM-Route) yields lower accuracy in both domains ($0.626$ and $0.744$). Providing the router with a playbook distilled from construction-split trajectories (LLM+Exp) improves its decisions ($0.655$ and $0.766$), but does not consistently outperform Exec-Lookup and incurs an additional routing call. Retrieving resources according to their semantic relevance to the task (ITR) achieves comparable accuracy ($0.670$ and $0.780$) but remains close to Full-$K_5$ in token usage ($0.99\times$ and $1.02\times$), compared with $0.86\times$ and $0.88\times$ for Exec-Lookup. Semantic relevance can therefore identify potentially useful resources, but does not directly determine a cost-efficient harness. Overall, measured class-level execution evidence provides the most consistent basis for initial harness selection.

\begin{figure*}[t]
  \centering
  \includegraphics[width=.9\textwidth]{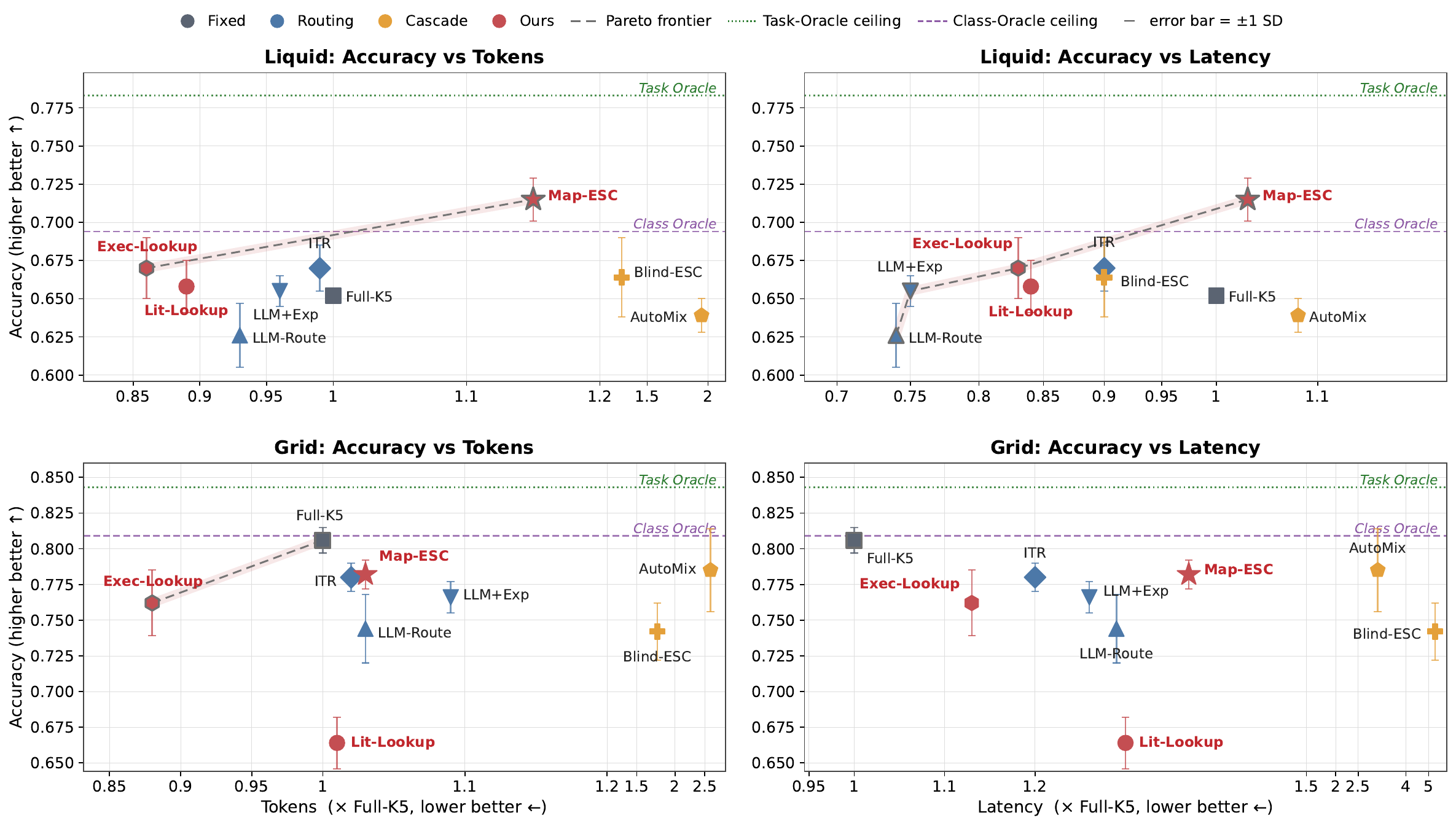} 
  \caption{Accuracy--Cost Trade-offs of Harness Provisioning Policies on the Test Split. Latency and tokens are normalized to Full-$K_5$; dashed lines show the empirical Pareto frontiers.}
\label{fig:pareto}
\end{figure*}

\subsection{RQ3: How Should the Map Guide Harness Selection?}
\label{sec:rq3}

The task-to-harness map supports two deployment modes: using the mapped harness directly or using it as the initial provision before conditional escalation. As shown in Table~\ref{tab:provisioning} and Figure~\ref{fig:pareto}, Exec-Lookup achieves accuracies of $0.670$ on \textsc{Liquid} and $0.762$ on \textsc{Grid}, while reducing token usage to $0.86\times$ and $0.88\times$ relative to Full-$K_5$. Map-ESC retries under $K_5$ when the initial execution fails its self-check, increasing accuracy to $0.715$ and $0.782$, with token costs of $1.15\times$ and $1.03\times$, respectively. Direct lookup therefore favors efficiency, while conditional escalation improves accuracy at additional cost.

Blind-ESC and AutoMix represent two alternative escalation strategies. Blind-ESC starts from $K_1$ and progressively expands the harness based on self-checks, whereas AutoMix uses a learned cascade to determine whether further provision is needed. Map-ESC instead selects the initial harness from the execution-derived map and allows at most one fallback to $K_5$. On \textsc{Liquid}, Map-ESC outperforms both Blind-ESC ($0.715$ versus $0.664$) and AutoMix ($0.715$ versus $0.639$). On \textsc{Grid}, it outperforms Blind-ESC ($0.782$ versus $0.742$) and nearly matches AutoMix ($0.782$ versus $0.785$), while using fewer tokens than both ($1.03\times$ versus $1.75\times$ and $2.61\times$). Map-guided two-stage escalation therefore provides the most consistent accuracy--efficiency trade-off across the two domains.

Figure~\ref{fig:pareto} further shows that choosing an appropriate harness configuration depends on both the domain and the deployment objective. On \textsc{Liquid}, Exec-Lookup provides a low-cost configuration, while Map-ESC achieves the highest accuracy among the harness-selection methods. On \textsc{Grid}, Full-$K_5$ remains the most accurate configuration, whereas Exec-Lookup offers a lower-cost Pareto-efficient alternative. The map therefore helps determine when additional provision improves performance and when it only increases cost.

Table~\ref{tab:execution_side} compares Map-ESC with execution-side methods operating under a fixed $K_5$ harness. Map-ESC matches Reflexion in accuracy ($0.715$ versus $0.711$) while using $48\%$ fewer tokens. ExpeL improves accuracy by $0.016$ but requires $44\%$ more tokens. Harness provisioning and execution-side adaptation are therefore complementary, with measured provisioning offering the stronger efficiency trade-off in this setting.\footnote{Appendix~\ref{app:additional_analysis} details routing-granularity results, escalation diagnostics, experience ablations, and qualitative error analysis.}

\begin{table}[h]
\centering
\caption{Provisioning Versus Execution-side Adaptation on the \textsc{Liquid} Test Split. Tokens are normalized to Map-ESC.}
\label{tab:execution_side}
\small
\setlength{\tabcolsep}{5pt}
\begin{tabular}{l l cc}
\toprule
Method & Adaptation lever & Acc.\ $\pm$ std & Tok. \\
\midrule
Map-ESC
& harness provisioning
& $\underline{0.715\pm0.014}$
& $\mathbf{1.00\times}$ \\

Reflexion
& retry and reflection
& $0.711\pm0.027$
& $1.91\times$ \\

ExpeL
& trajectory retrieval
& $\mathbf{0.731\pm0.009}$
& $1.44\times$ \\
\bottomrule
\end{tabular}
\end{table}

\subsection{RQ4: Does the Map Generalize?}
\label{sec:rq4}

The two domains exhibit the same qualitative need for task-aware provisioning but different deployment outcomes. As shown in Table~\ref{tab:provisioning} and Figure~\ref{fig:pareto}, task-aware harness selection improves both accuracy and cost in \textsc{Liquid}, whereas Full-$K_5$ remains accuracy-optimal in \textsc{Grid} and Exec-Lookup provides a lower-cost Pareto-efficient alternative. Task-dependent harness demand therefore persists across domains, although the benefit of reducing provision is domain-specific.

We further transfer the execution-derived map estimated with GPT-5.4 to Qwen3.5-27B without recalibration. As shown in Table~\ref{tab:cross_executor}, the transferred map improves Qwen over its Full-$K_5$ baseline on \textsc{Liquid} ($0.726$ versus $0.706$) and remains close on \textsc{Grid} ($0.797$ versus $0.808$). In contrast, a map estimated from a single Qwen run performs substantially worse in both domains. The task-to-harness relationship therefore transfers across executors, while reliable map estimation benefits from stronger and repeated measurements.

\begin{table}[h]
\centering
\caption{Cross-executor Transfer of the Execution-derived Map. All conditions use Qwen3.5-27B as the executor.}
\label{tab:cross_executor}
\small
\setlength{\tabcolsep}{5pt}
\begin{tabular}{l l cc}
\toprule
Policy & Map source & \textsc{Liquid} & \textsc{Grid} \\
\midrule
Full-$K_5$
& none
& $0.706$
& $\mathbf{0.808}$ \\

Exec-Lookup
& GPT-5.4
& $\mathbf{0.726}$
& $\underline{0.797}$ \\

Exec-Lookup
& Qwen
& $0.626$
& $0.762$ \\
\bottomrule
\end{tabular}
\end{table}

\section{Conclusion}

We formulate harness provisioning as an explicit and measurable deployment decision for LLM agents in mission-critical infrastructure. A task taxonomy derived from the mathematical formulation of physical systems characterizes task demand, while a cumulative $K_1$--$K_5$ hierarchy characterizes the information and capabilities available to the agent. We estimate the relationship between them using a literature-derived map of prior provisioning practices and an execution-derived map that selects the lowest harness level achieving near-best class-level performance. These maps guide either direct lookup or self-check-triggered escalation to $K_5$.
Our results show that suitable harness configurations vary across task classes and domains, and that additional provision may improve, leave unchanged, or reduce performance. 
They support selecting harness configurations according to task and domain requirements rather than granting maximal information, tools, and operational access by default.\\


\noindent\textbf{Limitations and Future Work.}
The measured task-to-harness relationship is calibrated to the evaluated executor, harness implementation, protocol, and tolerance. Its generalization to new domains, models, and harness implementations requires further validation and, where necessary, recalibration. Our evaluation covers 240 tasks in two simulation-backed environments, which do not fully capture live conditions such as sensor uncertainty, distribution shift, organizational procedures, and human approval. The literature-derived map may reflect publication and annotation bias, while the execution-derived map remains subject to finite-sample variation, judge error, and imperfect self-checks.
Future work will expand and release a broader benchmark for LLM agents in MCI O\&M, together with reproducible test environments covering additional domains, tasks, disturbances, and tool interfaces. We will also investigate learning-based methods that use the task-to-harness map to guide harness selection and escalation, and extend the evaluation framework to include security and operational risks.
\newpage
\bibliographystyle{ACM-Reference-Format}
\bibliography{sample-base}

@String{Computing = "Computing" }

@String{Chelsea = "Chelsea" }

@article{yang2026ToolPrivBench,
  title={When Lower Privileges Suffice: Investigating Over-Privileged Tool Selection in LLM Agents},
  author={Yang, Kaiyue and Bu, Yuyan and Yi, Jingwei and Wang, Yuchi and Zhou, Biyu and Dai, Juntao and Hu, Songlin and Yang, Yaodong},
  journal={arXiv preprint arXiv:2606.20023},
  year={2026}
}

@article{meng2026agentsurvey,
  title={Agent harness for large language model agents: A survey},
  author={Meng, Qianyu and Wang, Yanan and Chen, Liyi and Li, Yihang and Wu, Wei and Jiang, Wenyuan and Wang, Qimeng and Lu, Chengqiang and Gao, Yan and Wu, Yi and others},
  year={2026},
  publisher={Preprints}
}

@misc{li2026agentharness,
  title={Agent Harness Engineering: A Survey},
  author={Li, Junjie and Xiao, Xi and Zhang, Yunbei and Liu, Chen and
          Zhao, Lin and Liao, Xiaoying and Ji, Yingrui and Wang, Janet and
          Ge, Yingqiang and Xu, Weijie and Fang, Xi and Xu, Xiang and
          Zhao, Tianchen and Kim, Youngeun and Hamm, Jihun and
          Wang, Tianyang and Reddy, Chandan},
  url={https://openreview.net/pdf?id=eONq7FdiHa},
  year={2026}
}

@article{shi2025progent,
  title={Progent: Programmable privilege control for llm agents},
  author={Shi, Tianneng and He, Jingxuan and Wang, Zhun and Wu, Linyu and Li, Hongwei and Guo, Wenbo and Song, Dawn},
  journal={arXiv e-prints},
  pages={arXiv--2504},
  year={2025}
}

@article{kim2025prompt,
  title={Prompt flow integrity to prevent privilege escalation in llm agents},
  author={Kim, Juhee and Choi, Woohyuk and Lee, Byoungyoung},
  journal={arXiv preprint arXiv:2503.15547},
  year={2025}
}

@inproceedings{qin2024toolllm,
  title={Toolllm: Facilitating large language models to master 16000+ real-world apis},
  author={Qin, Yujia and Liang, Shihao and Ye, Yining and Zhu, Kunlun and Yan, Lan and Lu, Yaxi and Lin, Yankai and Cong, Xin and Tang, Xiangru and Qian, Bill and others},
  booktitle={International Conference on Learning Representations},
  volume={2024},
  pages={9695--9717},
  year={2024}
}

@article{mudunuri2026semantic,
  title={Semantic Tool Discovery for Large Language Models: A Vector-Based Approach to MCP Tool Selection},
  author={Mudunuri, Sarat and Wan, Jian and Qin, Ally and Manoharan, Srinivasan},
  journal={arXiv preprint arXiv:2603.20313},
  year={2026}
}

@inproceedings{jiang-etal-2023-active,
    title = "Active Retrieval Augmented Generation",
    author = "Jiang, Zhengbao  and
      Xu, Frank  and
      Gao, Luyu  and
      Sun, Zhiqing  and
      Liu, Qian  and
      Dwivedi-Yu, Jane  and
      Yang, Yiming  and
      Callan, Jamie  and
      Neubig, Graham",
    editor = "Bouamor, Houda  and
      Pino, Juan  and
      Bali, Kalika",
    booktitle = "Proceedings of the 2023 Conference on Empirical Methods in Natural Language Processing",
    month = dec,
    year = "2023",
    address = "Singapore",
    publisher = "Association for Computational Linguistics",
    url = "https://aclanthology.org/2023.emnlp-main.495/",
    doi = "10.18653/v1/2023.emnlp-main.495",
    pages = "7969--7992",
}

@inproceedings{asai2024self,
  title={Self-rag: Learning to retrieve, generate, and critique through self-reflection},
  author={Asai, Akari and Wu, Zeqiu and Wang, Yizhong and Sil, Avi and Hajishirzi, Hannaneh},
  booktitle={International conference on learning representations},
  volume={2024},
  pages={9112--9141},
  year={2024}
}

@inproceedings{joren2025sufficient,
  title={Sufficient context: A new lens on retrieval augmented generation systems},
  author={Joren, Hailey and Zhang, Jianyi and Ferng, Chun-Sung and Juan, Da-Cheng and Taly, Ankur and Rashtchian, Cyrus},
  booktitle={International Conference on Learning Representations},
  volume={2025},
  pages={20310--20334},
  year={2025}
}

@inproceedings{du2024anytool,
author = {Du, Yu and Wei, Fangyun and Zhang, Hongyang},
title = {AnyTool: self-reflective, hierarchical agents for large-scale API calls},
year = {2024},
publisher = {JMLR.org},
booktitle = {Proceedings of the 41st International Conference on Machine Learning},
articleno = {470},
numpages = {18},
location = {Vienna, Austria},
series = {ICML'24}
}

@inproceedings{lu2023chameleon,
author = {Lu, Pan and Peng, Baolin and Cheng, Hao and Galley, Michel and Chang, Kai-Wei and Wu, Ying Nian and Zhu, Song-Chun and Gao, Jianfeng},
title = {Chameleon: plug-and-play compositional reasoning with large language models},
year = {2023},
publisher = {Curran Associates Inc.},
address = {Red Hook, NY, USA},
booktitle = {Proceedings of the 37th International Conference on Neural Information Processing Systems},
articleno = {1882},
numpages = {32},
location = {New Orleans, LA, USA},
series = {NIPS '23}
}

@misc{ong2025routellm,
      title={RouteLLM: Learning to Route LLMs with Preference Data},
      author={Isaac Ong and Amjad Almahairi and Vincent Wu and Wei-Lin Chiang and Tianhao Wu and Joseph E. Gonzalez and M Waleed Kadous and Ion Stoica},
      year={2024},
      eprint={2406.18665},
      archivePrefix={arXiv},
      primaryClass={cs.LG},
      url={https://arxiv.org/abs/2406.18665},
}

@inproceedings{aggarwal2024automix,
author = {Aggarwal, Pranjal and Madaan, Aman and Anand, Ankit and Potharaju, Srividya Pranavi and Mishra, Swaroop and Zhou, Pei and Gupta, Aditya and Rajagopal, Dheeraj and Kappaganthu, Karthik and Yang, Yiming and Upadhyay, Shyam and Faruqui, Manaal and Mausam},
title = {AutoMix: automatically mixing language models},
year = {2024},
isbn = {9798331314385},
publisher = {Curran Associates Inc.},
address = {Red Hook, NY, USA},
booktitle = {Proceedings of the 38th International Conference on Neural Information Processing Systems},
articleno = {4164},
numpages = {35},
location = {Vancouver, BC, Canada},
series = {NIPS '24}
}

@inproceedings{yao2023react,
  title = {{ReAct}: Synergizing Reasoning and Acting in Language Models},
  author = {Yao, Shunyu and Zhao, Jeffrey and Yu, Dian and Du, Nan and Shafran, Izhak and Narasimhan, Karthik and Cao, Yuan},
  booktitle = {International Conference on Learning Representations (ICLR) },
  year = {2023},
  html = {https://arxiv.org/abs/2210.03629},
}

@inproceedings{shinn2023reflexion,
author = {Shinn, Noah and Cassano, Federico and Gopinath, Ashwin and Narasimhan, Karthik and Yao, Shunyu},
title = {Reflexion: language agents with verbal reinforcement learning},
year = {2023},
publisher = {Curran Associates Inc.},
address = {Red Hook, NY, USA},
booktitle = {Proceedings of the 37th International Conference on Neural Information Processing Systems},
articleno = {377},
numpages = {19},
location = {New Orleans, LA, USA},
series = {NIPS '23}
}

@inproceedings{zhao2024expel,
author = {Zhao, Andrew and Huang, Daniel and Xu, Quentin and Lin, Matthieu and Liu, Yong-Jin and Huang, Gao},
title = {ExpeL: LLM agents are experiential learners},
year = {2024},
isbn = {978-1-57735-887-9},
publisher = {AAAI Press},
url = {https://doi.org/10.1609/aaai.v38i17.29936},
doi = {10.1609/aaai.v38i17.29936},
booktitle = {Proceedings of the Thirty-Eighth AAAI Conference on Artificial Intelligence and Thirty-Sixth Conference on Innovative Applications of Artificial Intelligence and Fourteenth Symposium on Educational Advances in Artificial Intelligence},
articleno = {2188},
numpages = {11},
series = {AAAI'24/IAAI'24/EAAI'24}
}

@article{zhang2025agentsandbox,
  title={LLM Agents Should Employ Security Principles},
  author={Zhang, Kaiyuan and Su, Zian and Chen, Pin-Yu and Bertino, Elisa and Zhang, Xiangyu and Li, Ninghui},
  journal={arXiv preprint arXiv:2505.24019},
  year={2025}
}

@article{zhu2025miniscope,
  title={MiniScope: A Least Privilege Framework for Authorizing Tool Calling Agents},
  author={Zhu, Jinhao and Tseng, Kevin and Vernik, Gil and Huang, Xiao and Patil, Shishir G and Fang, Vivian and Popa, Raluca Ada},
  journal={arXiv preprint arXiv:2512.11147},
  year={2025}
}

@inproceedings{laptev2015egads,
		title={Generic and Scalable Framework for Automated Time-series Anomaly Detection},
		author={Laptev, Nikolay and Amizadeh, Saeed and Flint, Ian},
		booktitle={Proceedings of the 21th ACM SIGKDD International Conference on Knowledge Discovery and Data Mining},
		pages={1939--1947},
		year={2015},
		organization={ACM}
}

@inproceedings{ren2019timeseries,
author = {Ren, Hansheng and Xu, Bixiong and Wang, Yujing and Yi, Chao and Huang, Congrui and Kou, Xiaoyu and Xing, Tony and Yang, Mao and Tong, Jie and Zhang, Qi},
title = {Time-Series Anomaly Detection Service at Microsoft},
year = {2019},
isbn = {9781450362016},
publisher = {Association for Computing Machinery},
address = {New York, NY, USA},
url = {https://doi.org/10.1145/3292500.3330680},
doi = {10.1145/3292500.3330680},
booktitle = {Proceedings of the 25th ACM SIGKDD International Conference on Knowledge Discovery \& Data Mining},
pages = {3009–3017},
numpages = {9},
location = {Anchorage, AK, USA},
series = {KDD '19}
}

@inproceedings{li2022circa,
author = {Li, Mingjie and Li, Zeyan and Yin, Kanglin and Nie, Xiaohui and Zhang, Wenchi and Sui, Kaixin and Pei, Dan},
title = {Causal Inference-Based Root Cause Analysis for Online Service Systems with Intervention Recognition},
year = {2022},
isbn = {9781450393850},
publisher = {Association for Computing Machinery},
address = {New York, NY, USA},
url = {https://doi.org/10.1145/3534678.3539041},
doi = {10.1145/3534678.3539041},
booktitle = {Proceedings of the 28th ACM SIGKDD Conference on Knowledge Discovery and Data Mining},
pages = {3230–3240},
numpages = {11},
location = {Washington DC, USA},
series = {KDD '22}
}

@inproceedings{chen2025aiopslab,
author = {Ma, Minghua and Clark, Jackson and Zhang, Shenglin},
title = {AIOpsLab in Action: An Open Platform for AIOps Research},
year = {2025},
isbn = {9798400712760},
publisher = {Association for Computing Machinery},
address = {New York, NY, USA},
url = {https://doi.org/10.1145/3696630.3728619},
doi = {10.1145/3696630.3728619},
booktitle = {Proceedings of the 33rd ACM International Conference on the Foundations of Software Engineering},
pages = {1223–1227},
numpages = {5},
location = {Clarion Hotel Trondheim, Trondheim, Norway},
series = {FSE Companion '25}
}

@article{patel2025assetopsbench,
  title={Results and Retrospective Analysis of the CODS 2025 AssetOpsBench Challenge},
  author={Patel, Dhaval and Shyalika, Chathurangi and Yarrabothula, Suryanarayana Reddy and Yue, Ling and Lin, Shuxin and Zhou, Nianjun and Rayfield, James},
  journal={arXiv preprint arXiv:2605.08518},
  year={2026}
}

@inproceedings{hu2025agentgen,
author = {Hu, Mengkang and Zhao, Pu and Xu, Can and Sun, Qingfeng and Lou, Jian-Guang and Lin, Qingwei and Luo, Ping and Rajmohan, Saravan},
title = {AgentGen: Enhancing Planning Abilities for Large Language Model based Agent via Environment and Task Generation},
year = {2025},
isbn = {9798400712456},
publisher = {Association for Computing Machinery},
address = {New York, NY, USA},
url = {https://doi.org/10.1145/3690624.3709321},
doi = {10.1145/3690624.3709321},
booktitle = {Proceedings of the 31st ACM SIGKDD Conference on Knowledge Discovery and Data Mining V.1},
pages = {496–507},
numpages = {12},
location = {Toronto ON, Canada},
series = {KDD '25}
}

@inproceedings{ma2025toolmvr,
author = {Ma, Zhiyuan and Liu, Jiayu and Luo, Xianzhen and Huang, Zhenya and Zhu, Qingfu and Che, Wanxiang},
title = {Advancing Tool-Augmented Large Language Models via Meta-Verification and Reflection Learning},
year = {2025},
isbn = {9798400714542},
publisher = {Association for Computing Machinery},
address = {New York, NY, USA},
url = {https://doi.org/10.1145/3711896.3736835},
doi = {10.1145/3711896.3736835},
booktitle = {Proceedings of the 31st ACM SIGKDD Conference on Knowledge Discovery and Data Mining V.2},
pages = {2078–2089},
numpages = {12},
location = {Toronto ON, Canada},
series = {KDD '25}
}

@article{pan2026naturalharness,
  title={Natural-language agent harnesses},
  author={Pan, Linyue and Zou, Lexiao and Guo, Shuo and Ni, Jingchen and Zheng, Hai-Tao},
  journal={arXiv preprint arXiv:2603.25723},
  year={2026}
}

@article{lee2026metaharness,
  title={Meta-harness: End-to-end optimization of model harnesses},
  author={Lee, Yoonho and Nair, Roshen and Zhang, Qizheng and Lee, Kangwook and Khattab, Omar and Finn, Chelsea},
  journal={arXiv preprint arXiv:2603.28052},
  year={2026}
}

@article{yao2026harnessbench,
  title={Harness-Bench: Measuring harness effects across models in realistic agent workflows},
  author={Yao, Yilun and Tan, Xinyu and Liu, Chao-Hsuan and Li, Yaoming and Wang, Zhengyang and Yu, Wenhan and Tan, Zhewen and Tian, Yuxuan and Zhao, Guangxiang and Sun, Lin and others},
  journal={arXiv preprint arXiv:2605.27922},
  year={2026}
}

@article{liu2026harnessaudit,
  title={Auditing agent harness safety},
  author={Liu, Chengzhi and Guo, Yichen and Liu, Yepeng and Yang, Yuzhe and Yan, Qianqi and Zhao, Xuandong and Hua, Wenyue and Liu, Sheng and Li, Sharon and Bu, Yuheng and others},
  journal={arXiv preprint arXiv:2605.14271},
  year={2026}
}

@inproceedings{su2019omnianomaly,
author = {Su, Ya and Zhao, Youjian and Niu, Chenhao and Liu, Rong and Sun, Wei and Pei, Dan},
title = {Robust Anomaly Detection for Multivariate Time Series through Stochastic Recurrent Neural Network},
year = {2019},
isbn = {9781450362016},
publisher = {Association for Computing Machinery},
address = {New York, NY, USA},
url = {https://doi.org/10.1145/3292500.3330672},
doi = {10.1145/3292500.3330672},
booktitle = {Proceedings of the 25th ACM SIGKDD International Conference on Knowledge Discovery \& Data Mining},
pages = {2828–2837},
numpages = {10},
location = {Anchorage, AK, USA},
series = {KDD '19}
}

@inproceedings{lai2025llmlight,
author = {Lai, Siqi and Xu, Zhao and Zhang, Weijia and Liu, Hao and Xiong, Hui},
title = {LLMLight: Large Language Models as Traffic Signal Control Agents},
year = {2025},
isbn = {9798400712456},
publisher = {Association for Computing Machinery},
address = {New York, NY, USA},
url = {https://doi.org/10.1145/3690624.3709379},
doi = {10.1145/3690624.3709379},
booktitle = {Proceedings of the 31st ACM SIGKDD Conference on Knowledge Discovery and Data Mining V.1},
pages = {2335–2346},
numpages = {12},
location = {Toronto ON, Canada},
series = {KDD '25}
}

@inproceedings{mohammadi2025evaluation,
author = {Mohammadi, Mahmoud and Li, Yipeng and Lo, Jane and Yip, Wendy},
title = {Evaluation and Benchmarking of LLM Agents: A Survey},
year = {2025},
isbn = {9798400714542},
publisher = {Association for Computing Machinery},
address = {New York, NY, USA},
url = {https://doi.org/10.1145/3711896.3736570},
doi = {10.1145/3711896.3736570},
booktitle = {Proceedings of the 31st ACM SIGKDD Conference on Knowledge Discovery and Data Mining V.2},
pages = {6129–6139},
numpages = {11},
location = {Toronto ON, Canada},
series = {KDD '25}
}

@article{franko2025dynamic,
  title={Dynamic system instructions and tool exposure for efficient agentic LLMs},
  author={Franko, Uria},
  journal={arXiv preprint arXiv:2602.17046},
  year={2025}
}

@misc{openai2026gpt54,
  author       = {{OpenAI}},
  title        = {{Introducing GPT‑5.4}},
  year         = {2026},
  month        = march,
  howpublished = {\url{https://openai.com/index/introducing-gpt-5-4/}},
}

@misc{googledeepmind2026gemini31pro,
  author       = {{Google DeepMind}},
  title        = {{Gemini 3.1 Pro: Model Card}},
  year         = {2026},
  month        = feb,
  howpublished = {\url{https://deepmind.google/models/model-cards/gemini-3-1-pro/}},
}

@article{yu2026skill,
  title={Skill is Not One-Size-Fits-All: Model-Aware Skill Alignment for LLM Agents},
  author={Yu, Jianxiang and Zhu, Jiapeng and Lin, Bochen and Cui, Qier and Ding, Zichen and Li, Xiang},
  journal={arXiv preprint arXiv:2605.30723},
  year={2026}
}

@article{endsley1995toward,
  title={Toward a Theory of Situation Awareness in Dynamic Systems},
  author={Endsley, Mica R},
  journal={Human Factors: The Journal of the Human Factors and Ergonomics Society},
  volume={37},
  number={1},
  pages={32--64},
  year={1995},
  publisher={SAGE Publications}
}

@article{rasmussen1985role,
  title={The role of hierarchical knowledge representation in decisionmaking and system management},
  author={Rasmussen, Jens},
  journal={IEEE Transactions on systems, man, and cybernetics},
  number={2},
  pages={234--243},
  year={1985},
  publisher={IEEE}
}

@misc{cisa_critical_infrastructure,
  author       = {{Cybersecurity and Infrastructure Security Agency}},
  title        = {Critical Infrastructure Security and Resilience},
  howpublished = {\url{https://www.cisa.gov/topics/critical-infrastructure-security-and-resilience}},
}

@ARTICLE{pandapower.2018,
author={L. Thurner and A. Scheidler and F. Schafer and J. H. Menke and J. Dollichon and F. Meier and S. Meinecke and M. Braun},
journal={IEEE Transactions on Power Systems},
title={pandapower - an Open Source Python Tool for Convenient Modeling, Analysis and Optimization of Electric Power Systems},
year={2018},
doi={10.1109/TPWRS.2018.2829021},
url={https://arxiv.org/abs/1709.06743},
ISSN={0885-8950}
}

@inproceedings{marot2021learning,
  title={Learning to run a power network challenge: a retrospective analysis},
  author={Marot, Antoine and Donnot, Benjamin and Dulac-Arnold, Gabriel and Kelly, Adrian and O’Sullivan, Aidan and Viebahn, Jan and Awad, Mariette and Guyon, Isabelle and Panciatici, Patrick and Romero, Camilo},
  booktitle={NeurIPS 2020 competition and demonstration track},
  pages={112--132},
  year={2021},
  organization={PMLR}
}

\appendix

\section{Literature Corpus and Annotation}
\label{app:literature_analysis}

\subsection{Corpus Collection and Screening}
\begin{figure*}[t]
  \centering
  \includegraphics[height=4.6cm]{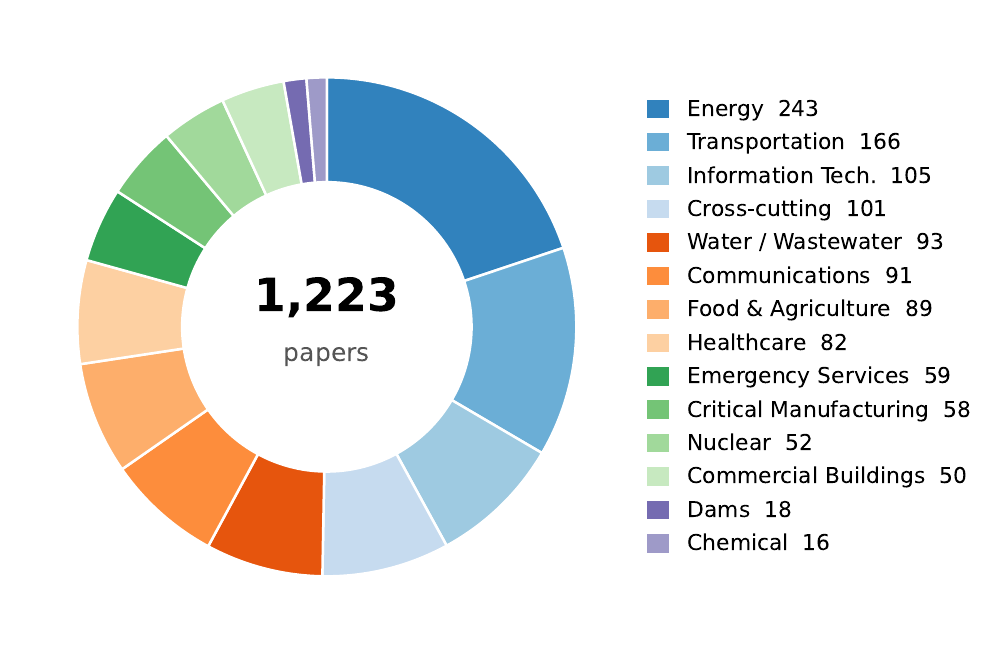}\hfill
  \includegraphics[height=4.6cm]{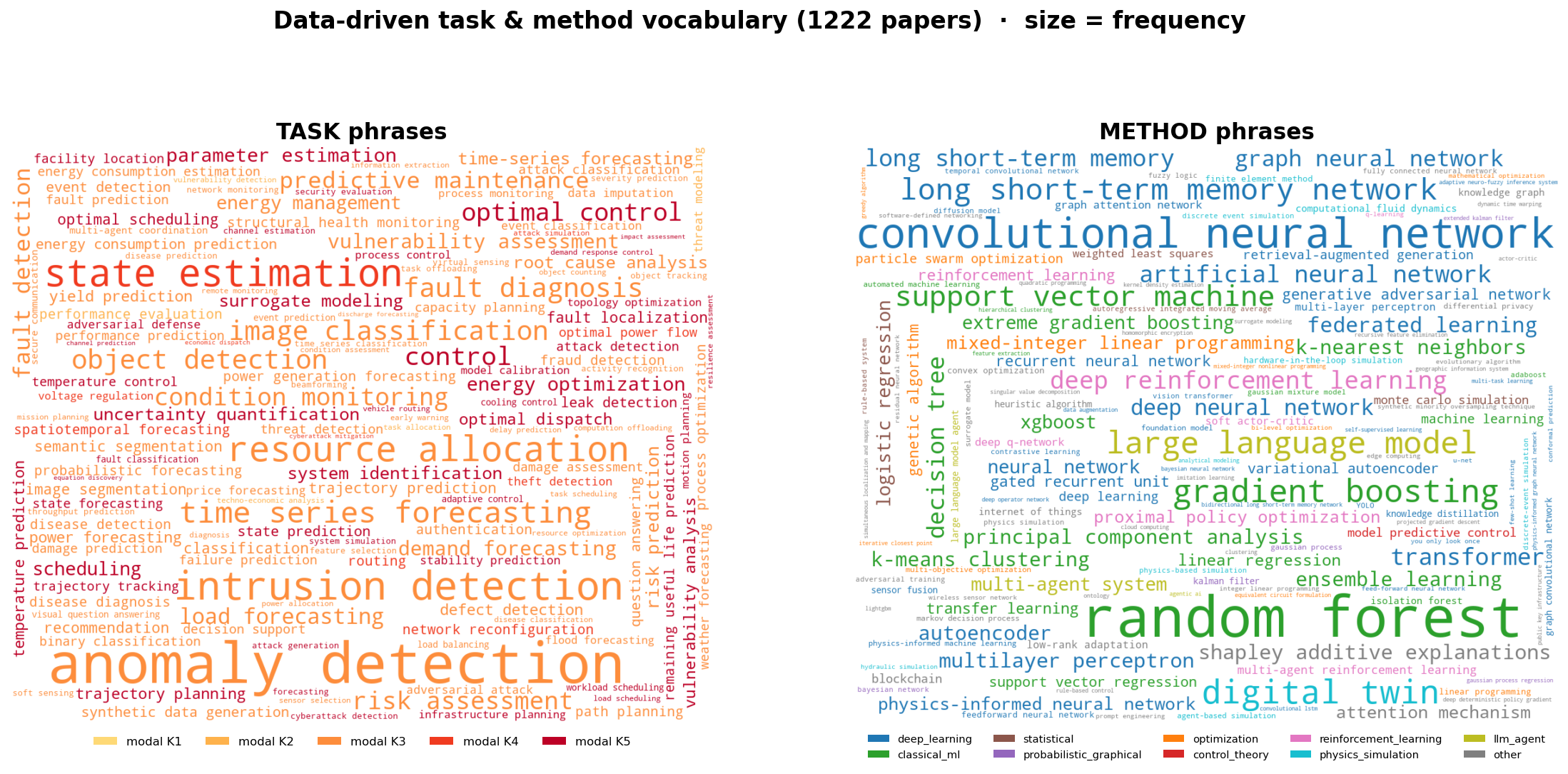}
  \caption{The literature corpus. Left: sector distribution. Right: data-driven task and method vocabulary; word size is corpus frequency.}
  \label{fig:lit_corpus}
\end{figure*}

We survey how the last decade (2016--2026) of MCI operation-and-maintenance research provisions system access, covering 13 of the 16 CISA critical-infrastructure sectors\footnote{\url{https://www.cisa.gov/topics/critical-infrastructure-security-and-resilience/critical-infrastructure-sectors}} (Defense Industrial Base, Financial Services, and Government Facilities are out of scope). Candidates come from three public indices, arXiv, Semantic Scholar, and OpenAlex, queried with sector vocabulary only (``substation'', ``chiller plant'', \dots): no task or method words, so the task and harness distributions emerge from the papers rather than from the query set. The collection and screening pipeline was built and operated with Claude Code (Fable 5). From roughly 2{,}000 crawled candidates, an LLM screening agent triages title and abstract, discarding surveys, position papers, and papers without a primary MCI O\&M task, and full-text extraction runs on the survivors. The screened corpus contains 1,223 papers, of which 1,220 receive complete task-coordinate and harness-level annotations and are used to estimate the literature-derived map; Figure~\ref{fig:lit_corpus} shows its sector distribution and vocabulary.

\subsection{Task and Harness Annotation Schema}
\begin{figure}[t]
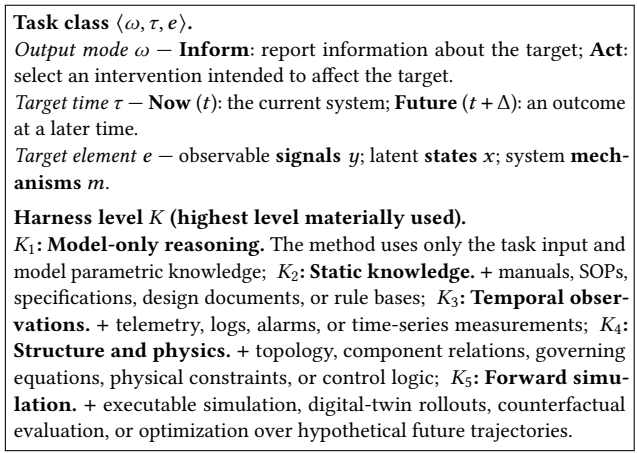

\centering
\fbox{\parbox{0.955\linewidth}{\small
\textbf{Task class $\langle\omega,\tau,e\rangle$.}\\
\emph{Output mode} $\omega$ --- \textbf{Inform}: report information about the target; \textbf{Act}: select an intervention intended to affect the target.\\
\emph{Target time} $\tau$ --- \textbf{Now} ($t$): the current system; \textbf{Future} ($t+\Delta$): an outcome at a later time.\\
\emph{Target element} $e$ --- observable \textbf{signals} $y$; latent \textbf{states} $x$; system \textbf{mechanisms} $m$.\\[3.5pt]
\textbf{Harness level $K$ (highest level materially used).}\\
\textbf{$K_1$: Model-only reasoning.} The method uses only the task input and model parametric knowledge;\;
\textbf{$K_2$: Static knowledge.} $+$ manuals, SOPs, specifications, design documents, or rule bases;\;
\textbf{$K_3$: Temporal observations.} $+$ telemetry, logs, alarms, or time-series measurements;\;
\textbf{$K_4$: Structure and physics.} $+$ topology, component relations, governing equations, physical constraints, or control logic;\;
\textbf{$K_5$: Forward simulation.} $+$ executable simulation, digital-twin rollouts, counterfactual evaluation, or optimization over hypothetical future trajectories.}}
\caption{Abridged annotation rubric aligned with the task and harness spaces defined in Section~\ref{sec:framework}.}
\label{box:rubric}
\end{figure}

Each paper receives two annotations from separate model calls that never see each other's output, a task coordinate and a harness level, so the class label cannot leak into the harness label. The task call is method- and name-blind: it returns the paper's primary task as one method-free sentence plus only the three axis labels of the taxonomy; no class names appear anywhere in the prompt. The harness call assigns the highest level the proposed solution \emph{materially} uses, considering both what the deployed method consumes at run time and what building it required, a policy trained in a simulator is K5 even if inference reads only telemetry. Figure~\ref{box:rubric} summarizes both rubrics in the paper's vocabulary.

\subsection{Multi-LLM Annotation Protocol}

We use three annotators from different model providers, including \texttt{gemini-3.1-pro}, \texttt{gpt-5.4-mini}, and \texttt{glm-5.2}, to independently assign each paper a task coordinate $\langle\omega,\tau,e\rangle$ and a harness level $K$, together with supporting evidence from the paper. The task axes are consolidated separately by majority vote, while the harness level is consolidated as a single ordinal label. Majority voting directly resolves 89\% of task-axis labels and 97\% of harness labels. The remaining cases, comprising 30 task-coordinate assignments and 36 harness-level assignments, undergo a separate adjudication pass assisted by Claude Fable 5. The adjudicator reviews the paper evidence without access to the original model votes, and the resulting labels are manually verified. Table~\ref{tab:lit_agreement} reports the consensus composition and inter-annotator agreement.

\begin{figure}[t]
  \centering
  \includegraphics[width=\linewidth]{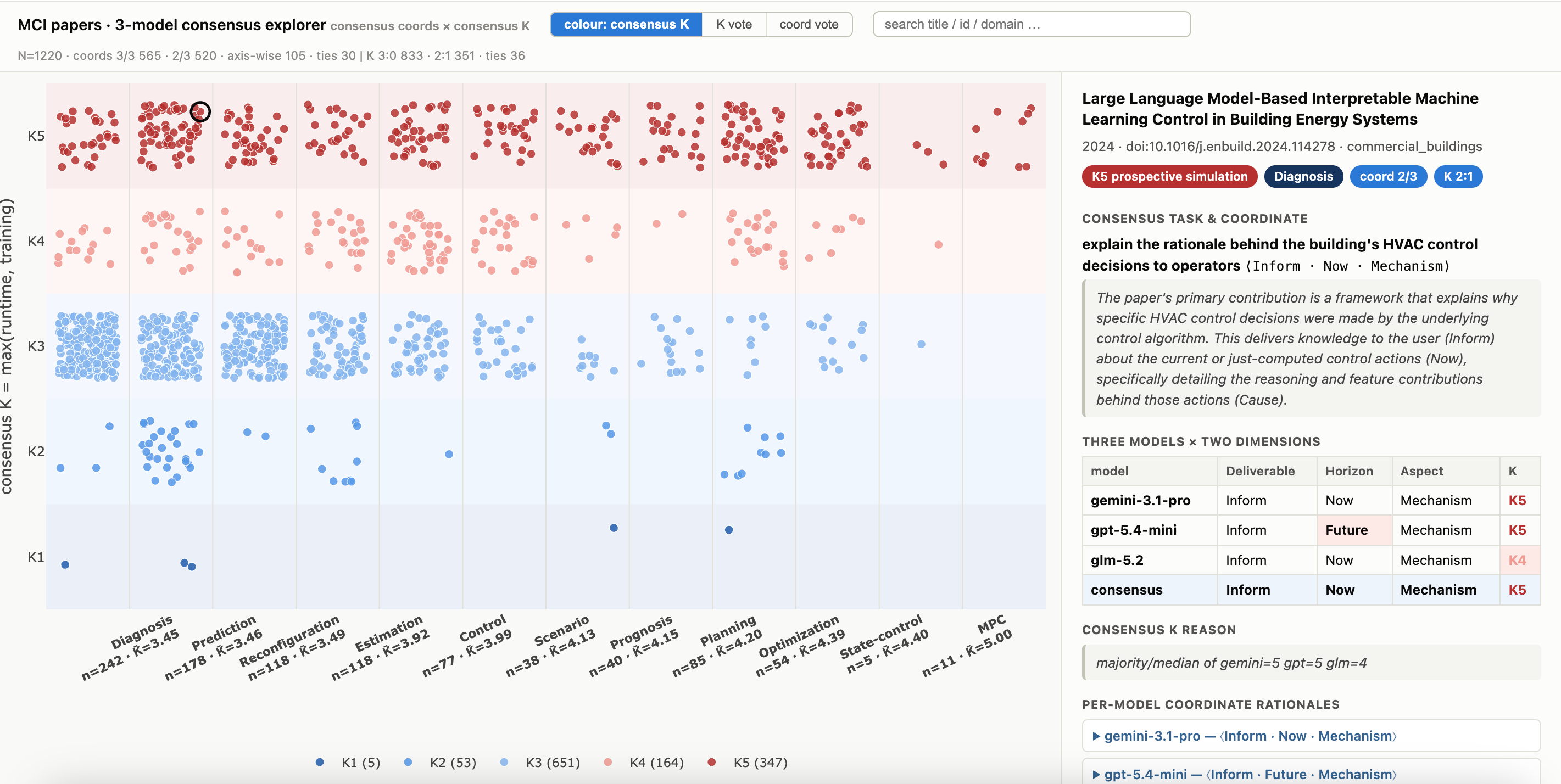}
  \caption{\textbf{Snapshot of the interactive literature explorer to be released with the project.} Each point represents one paper, grouped by task class and consensus harness level. The inspection panel presents the selected paper's metadata, supporting evidence, individual annotations, and consolidated labels.}
  \label{fig:lit_explorer}
\end{figure}

We will release the complete extraction through an interactive literature explorer, including each paper's task coordinate, harness level, supporting evidence, individual annotations, and consolidated labels. The explorer also supports inspection and filtering across task classes and harness levels; Figure~\ref{fig:lit_explorer} shows a snapshot.

\begin{table}[t]
\caption{Consensus composition and inter-annotator agreement.}
\label{tab:lit_agreement}
\centering
\small
\setlength{\tabcolsep}{3pt}
\begin{tabular}{lcccl}
\toprule
& 3:0 & 2:1 & adjudicated & pairwise agreement \\
\midrule
Task coordinate & 46\% & 54\% & 30 papers & $\kappa$ 0.49--0.84 by axis \\
Harness level   & 68\% & 29\% & 36 papers & exact 73--84\%; QWK 0.61--0.78 \\
\bottomrule
\end{tabular}
\end{table}

\subsection{Corpus-Level Findings}
\label{app:lit_stats}

Figure~\ref{fig:lit_dist} shows the harness distribution for each task class, while Table~\ref{tab:lit_full} reports the counts used to estimate $P_{\mathrm{lit}}(K\mid T)$. Each task axis is associated with higher harness provision: the mean level increases by $0.45$ from Inform to Act and by $0.39$ from Now to Future, while the marginal mean rises from $3.53$ for observable signals to $3.65$ for system mechanisms and $4.06$ for latent states. Provision is concentrated at $K_3$ and $K_5$, which account for $53\%$ and $28\%$ of the corpus, respectively, compared with $13\%$ at $K_4$.

\begin{figure}[h]
  \centering
  \includegraphics[width=0.8\linewidth]{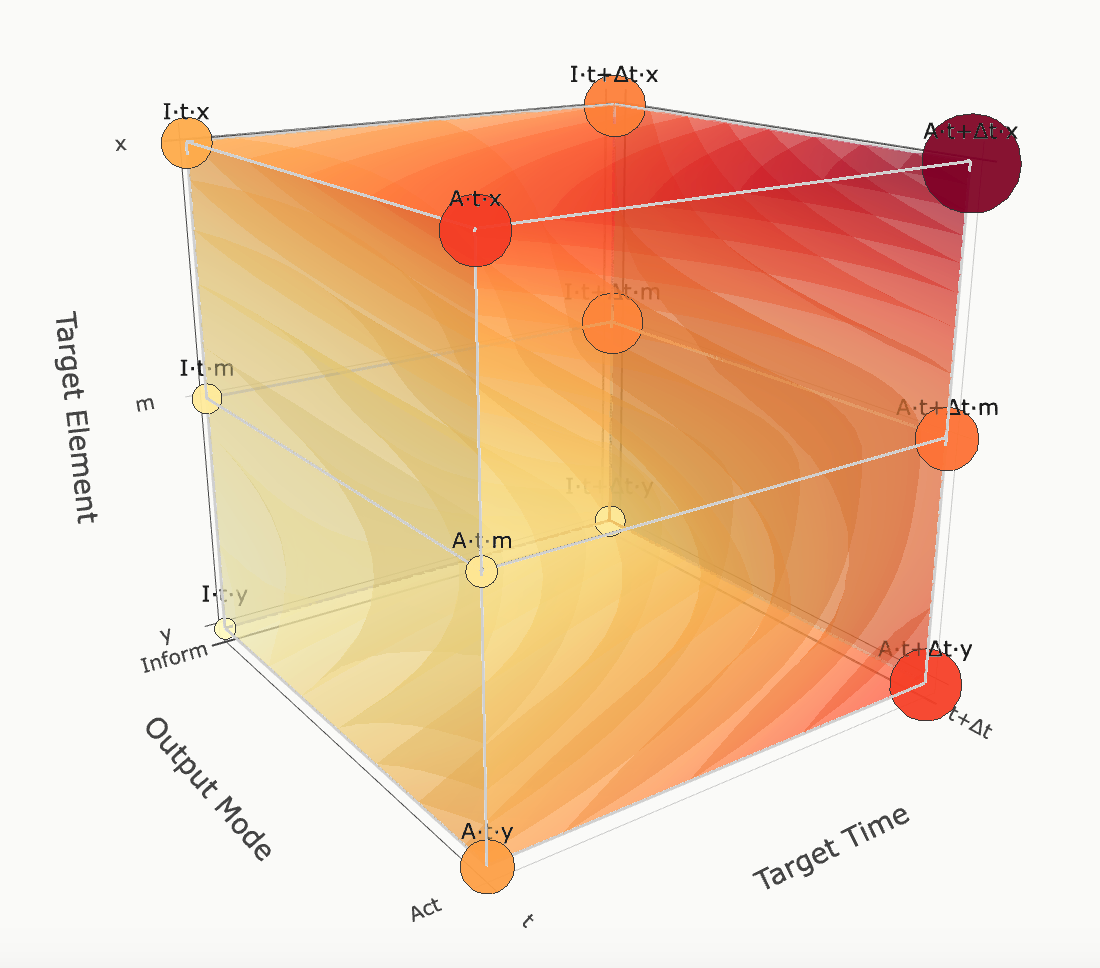}
    \caption{\textbf{Three-dimensional view of the literature-derived harness landscape.}
    The vertices represent the 12 task classes defined by output mode, target time, and target element, and surface colour indicates the mean literature-derived harness level $\bar{K}$. Conventional task names are shown only as approximate interpretations of the formal coordinates.}
    \label{fig:lit_3d_landscape}
\end{figure}

Figure~\ref{fig:lit_3d_landscape} provides a joint view of these trends over the complete $2\times2\times3$ task space. Harness provision generally increases toward Act, Future, and latent-state tasks, with the highest values concentrated around A-F-$x$ and adjacent Future--Act classes. The variation is nevertheless not determined by any single axis: task classes sharing one or two coordinates can still exhibit different literature-derived harness distributions.

The corpus records the harness level used by prior methods rather than the level sufficient for a fixed agent and environment. Estimates for sparsely represented classes, particularly A-N-$x$ ($n=5$) and A-F-$x$ ($n=11$), are also less reliable. We therefore treat the literature-derived map as a prior rather than verified execution evidence. More broadly, the analysis provides a survey-scale account of how MCI O\&M research has provisioned system information and
capabilities across task classes.

\begin{table}[t]
\caption{Literature-derived harness distributions across the 12 task classes, based on the consensus labels of three annotators. Counts are reported by harness level, and task classes are ordered by mean $K$.}
\label{tab:lit_full}
\centering
\small
\setlength{\tabcolsep}{3.4pt}
\begin{tabular}{lrrrrrrcc}
\toprule
Task class $\langle\omega,\tau,e\rangle$
& $n$ & $K_1$ & $K_2$ & $K_3$ & $K_4$ & $K_5$
& $\bar{K}$ & Mode \\
\midrule
I-N-$y$ & 254 & 1 & 3  & 208 & 13 & 29 & 3.26 & $K_3$ \\
I-N-$m$ & 242 & 2 & 27 & 132 & 22 & 59 & 3.45 & $K_3$ \\
I-F-$y$ & 178 & 0 & 2  & 127 & 14 & 35 & 3.46 & $K_3$ \\
A-N-$m$ & 118 & 0 & 9  & 65  & 21 & 23 & 3.49 & $K_3$ \\
I-N-$x$ & 118 & 0 & 1  & 44  & 36 & 37 & 3.92 & $K_3$ \\
A-N-$y$ & 77  & 0 & 0  & 28  & 22 & 27 & 3.99 & $K_3$ \\
I-F-$m$ & 38  & 1 & 2  & 9   & 5  & 21 & 4.13 & $K_5$ \\
I-F-$x$ & 40  & 0 & 0  & 16  & 2  & 22 & 4.15 & $K_5$ \\
A-F-$m$ & 85  & 1 & 9  & 8   & 21 & 46 & 4.20 & $K_5$ \\
A-F-$y$ & 54  & 0 & 0  & 13  & 7  & 34 & 4.39 & $K_5$ \\
A-N-$x$ & 5   & 0 & 0  & 1   & 1  & 3  & 4.40 & $K_5$ \\
A-F-$x$ & 11  & 0 & 0  & 0   & 0  & 11 & 5.00 & $K_5$ \\
\bottomrule
\end{tabular}
\end{table}

\section{Benchmark Construction and Verification}
\label{app:benchmark}

\subsection{Simulation Environments}
\textbf{Liquid.} A differentiable thermal-hydraulic digital twin of a 10-rack liquid-cooled data hall: one CDU loop (pump, control valve, facility heat exchanger) feeding 20 server cold plates over a 130-node/149-edge hydraulic network. Transient states integrate with \texttt{dopri5} at $\Delta t{=}5$\,s (rtol $10^{-4}$, atol $10^{-6}$); steady states solve the coupled hydraulic--thermal system directly. An episode spans 0--1200\,s. Telemetry is a $\sim$910-column table of \texttt{family:target} signals (node temperatures and pressures, flows, pump speed, per-plate power); the internal ODE states (e.g., metal base-plate temperatures) are recorded but hidden from the agent, which grounds the latent-state ($x$) task classes.

\textbf{Grid.} A power-grid environment on Grid2Op's \path{l2rpn_case14_sandbox} (IEEE-14: 14 substations, 20 branches, 6 generators, 11 loads), with pandapower AC power flow as the K5 engine. Reference trajectories are 288-step days at 5-minute resolution with overflow disconnection disabled so that overloads remain observable. Telemetry is a 142-column table (per-line loading $\rho$, flows, currents, voltages, status; per-generator and per-load injections). Here latent-state tasks arise from \emph{structurally absent} columns---network losses and the full bus-voltage vector are in no column, and three buses have no voltage sensor---rather than from masking.

\subsection{Scenario Generation}
Liquid uses six simulated trajectories (two normal-operation trajectories, single- and multi-plate overheating, a migrating hotspot, and a pump degradation ramp). Grid uses ten one-day windows selected by a two-pass scan over 1{,}004 Grid2Op chronics: two normal days, seven sustained-overload windows spanning three distinct bottleneck lines at increasing severity ($\rho_{\max}$ 1.00--1.26), and one mid-window line-trip fault that cascades to $\rho{=}1.95$ on a neighboring line. Every task binds a scenario file, an observation window, and a current time; the visibility gate (Appendix~\ref{app:harness_implementation}) prevents future-peeking, so scenarios are replayable and deterministic.

\subsection{Task Construction}
Each domain instantiates 10 tasks for each of the 12 classes (240 total) from parameterized template families (e.g., \path{value_at}, \path{hidden_plate_temp}, \path{best_action_rho}, \path{min_pump_below_maxtemp}), with natural-language paraphrase applied on top of templates. Every task carries a machine-readable ground-truth specification \path{gt_spec}=\{method, args\} resolved by a deterministic oracle ($\sim$80 methods for Liquid, 54 for Grid) with three evidence tiers: direct trajectory lookups, closed-form physics (e.g., $Q=\rho\dot V c_p\Delta T$), and simulator co-solves that call the \emph{same} engine exposed to the agent at K5, with privileged access (full trajectory, no visibility gate, hidden states). Benchmark templates are written against the system representation, not against any literature paper, so the literature map and the benchmark share only the taxonomy.

\subsection{Construction and Test Splits}
Each class splits 5/5 into a construction (map-estimation) split and a held-out test split, stratified by template family so no family appears only in one split, with a fixed RNG seed. The full $K_1$--$K_5$ sweep runs on all 240 tasks, but execution-derived levels are estimated \emph{only} from construction tasks and all policies are evaluated \emph{only} on test tasks.

\subsection{Programmatic and Human Verification}
Programmatic checks: (1)~the ground truth, re-fed verbatim to the grader, must score 1.0 on all 240 tasks; (2)~a cross-split echo audit verifies that no test answer string appears in construction traces; (3)~independent oracle re-derivation validates every stored ground truth; (4)~a probe pass removes degenerate axes discovered during construction (e.g., the slack generator's inert setpoint and a voltage-pinned bus in Grid, which would otherwise make some ``counterfactual'' tasks trivial or dead). Human review of sampled traces additionally corrected two Grid ground-truth values after an audit of the prediction/prognosis cells; the affected 20 tasks were re-executed at all levels before any map or policy result reported here.

\subsection{Representative Tasks}
Table~\ref{tab:task_examples} shows one test task per class from the Liquid domain (Grid analogues replace temperatures with line loadings and pump/valve actions with generation redispatch).

\begin{table*}[t]
\caption{Representative Liquid benchmark tasks, one per class. S/X/M denote signal($y$)/state($x$)/mechanism targets($m$); graders are deterministic rule checks combined with an LLM judge (Appendix~\ref{app:experimental_details}).}
\label{tab:task_examples}
\centering
\small
\begin{tabular}{llp{9.2cm}l}
\toprule
Class & Task & Question (abridged) & Grading \\
\midrule
I-N-$y$ & I-N-S-01 & What is the outlet water temperature at cold plate \texttt{rack\_a4\_s1} right now? & numeric, $\pm$0.6\,K \\
I-N-$x$ & I-N-X-01 & How hot is the (unsensored) metal base plate of \texttt{rack\_a4\_s1} right now? & numeric, $\pm$1.5\,K \\
I-N-$m$ & I-N-M-02 & Outlets run high on both sides of the hall; which single device is operating far outside its typical range? & set match \\
I-F-$y$ & I-F-S-01 & The return-header temperature is drifting; using data up to now, where will it be at $t{=}650$\,s? & numeric, $\pm$0.4\,K \\
I-F-$x$ & I-F-X-02 & \texttt{rack\_a3\_s2} steps from 503\,W to 1400\,W at $t{\approx}305$\,s; what will its base-plate temperature reach? & numeric, $\pm$1.5\,K \\
I-F-$m$ & I-F-M-03 & If the pump were raised to 4500\,rpm at today's load, where would the hottest outlet settle? & numeric, $\pm$0.8\,K \\
A-N-$y$ & A-N-S-02 & SLA: all outlets $\le$300\,K. Does the hall need an intervention right now? & boolean \\
A-N-$x$ & A-N-X-02 & Policy: every base plate $\le$310\,K (not directly measurable). Is an intervention needed now? & boolean \\
A-N-$m$ & A-N-M-01 & One chip is overheating; which cold plate do you act on? & set match \\
A-F-$y$ & A-F-S-01 & Lowest pump speed in \{3000..6000\}\,rpm keeping the hottest outlet within 305\,K? & numeric, $\pm$1\,rpm \\
A-F-$x$ & A-F-X-03 & If the pump goes to 5500\,rpm now, how much lower is \texttt{rack\_a3\_s2}'s base plate a minute later vs.\ no action? & numeric, $\pm$0.6\,K \\
A-F-$m$ & A-F-M-01 & Choose one measure to cool the hall: raise pump to 3500\,rpm, migrate half of one rack's load, or open the valve. & set match \\
\bottomrule
\end{tabular}
\end{table*}

\section{Harness Implementations}
\label{app:harness_implementation}

\subsection{Cumulative Harness Manifests}
Levels are strictly cumulative ($\mathcal H_{K_1}\subset\cdots\subset\mathcal H_{K_5}$): each level's registry unions the previous level's tools, and an agent at level $K$ is constructed with exactly that registry. Table~\ref{tab:harness_manifests} lists both manifests.

\begin{table*}[t]
\caption{Resources exposed at each harness level. Tools are OpenAI function-calling interfaces; each level unions all lower levels.}
\label{tab:harness_manifests}
\centering
\small
\begin{tabular}{lp{6.6cm}p{6.6cm}}
\toprule
Level & Liquid & Grid \\
\midrule
K1 model-only & no tools; parametric reasoning over the task prompt & no tools; parametric reasoning over the task prompt \\
K2 static & \texttt{static\_knowledge} (catalog, parameter schemas, topology summary, typical ranges, units); \texttt{web\_search} & \texttt{static\_knowledge}; \texttt{search\_knowledge} and \texttt{search\_sop} (RAG over references and operator playbooks); \texttt{web\_search} \\
K3 temporal & \texttt{list\_signals}, \texttt{get\_signal\_window}, \texttt{compute\_signal\_stats}, \texttt{find\_events} over gated telemetry & same four tools over the 142-column grid trajectory \\
K4 structure/physics & \texttt{query\_topology}, \texttt{get\_fluid\_properties}, \texttt{check\_physics} (energy balance, pressure drop, mixing), \texttt{check\_constraints} & \texttt{query\_topology}, \texttt{get\_component\_limits}, \texttt{check\_physics} (power balance, loading, per-unit V), \texttt{check\_constraints} \\
K5 simulation & \texttt{simulate\_steady} (hydraulic/coupled), \texttt{what\_if}, \texttt{compute\_sensitivity}, \texttt{simulate\_forward} (transient rollout) & \texttt{run\_powerflow} (AC, with overrides), \texttt{what\_if}, \texttt{contingency\_analysis} (N-1), \texttt{compute\_sensitivity} \\
\bottomrule
\end{tabular}
\end{table*}

\subsection{Tool and Data Interfaces}
All tools are exposed as JSON function-calling schemas; results return as structured JSON (time series are capped at 100 points per call, transient rollouts at 18 steps). K5 override conventions are uniform (pump speed, valve opening, per-plate power in Liquid; generator/load MW and line outages in Grid). Simulator wall-time and call counts are metered separately per task.

\subsection{Capability Enforcement}
Capability boundaries are architectural, not prompt-level: a level-$K$ agent's process is constructed with only that level's tool registry, so higher-level tools are not merely discouraged but nonexistent in its API schema. A shared visibility gate clamps every data access to the task's observation horizon (agent-supplied times are clamped server-side, including K5 rollout start times), and hidden state families are masked at the data source; the grading oracle deliberately bypasses both. During escalation, only the model's own textual findings are carried into the next attempt---no tool outputs, caches, or context cross the boundary.

\subsection{Domain-Specific Differences}
K1--K3 are semantically identical across domains (same tool names, same \texttt{kind:target} signal convention, same agent shell). K4 differs by physics content only (hydraulic network and heat balances vs.\ electrical topology, thermal limits, and power balances). K5 differs structurally: Liquid exposes a transient rollout because minute-scale thermal dynamics are genuine physics (plate thermal mass, exchanger lag), whereas the grid is quasi-static at 5-minute resolution---given injections, the state follows algebraically from power flow---so steady AC power flow plus N-1 screening \emph{is} the complete simulation capability. A grid time-rollout is deliberately not exposed: the grid's future is driven by exogenous injection trajectories, so stepping the environment forward would either replay recorded future injections (leaking ground truth through the visibility gate) or reduce to the already-provided counterfactual solve under hypothesized injections; extrapolating those injections from telemetry is the agent's own K3-level task, not a harness capability. Latent state is gated telemetry in Liquid but structural sensor absence in Grid---two realistic mechanisms for the same taxonomy coordinate.


\section{Execution and Evaluation Protocol}
\label{app:experimental_details}

\subsection{Executor Configuration}
All main results use a frozen \texttt{gpt-5.4} executor behind a ReAct-style function-calling loop: one model call per iteration, emitted tool calls executed and returned, until the model answers without tool calls or hits the iteration cap (20). Sampling: temperature 0.2, no output-token cap. Each task runs in an isolated spawned process with a 300\,s hard timeout and a soft deadline 45\,s earlier; on exhaustion the agent receives one final tool-free ``answer from the evidence you have'' call, so timeouts degrade to a best-effort answer rather than an empty one. The cross-executor study serves \texttt{Qwen3.5-27B} on a single-node vLLM server (thinking disabled, native tool-call parser, server-side prompt-tail truncation to fit the context window) behind the identical loop, harness, and tasks.

\subsection{Task Scoring}
Each task is scored in $[0,1]$ as the mean of a deterministic rule score and an LLM-judge score. Rule graders are typed per task: numeric-with-tolerance (key-anchored extraction with unit normalization), set match (Jaccard), boolean polarity, directional, and required-fact containment. Inform tasks grade the reported estimate against the oracle value; Act tasks grade the selected intervention against the simulator-verified best action (e.g., the cheapest pump setting that satisfies the constraint), so action quality is grounded in the same physics the oracle uses. 

\subsection{LLM Judge}
The judge is \texttt{gemini-3.1-pro} at temperature 0 with a fixed rubric (1.0 fully correct $\to$ 0.0 wrong or empty; credit substance over phrasing and unit differences). It sees only the question, ground truth, and final answer---never the harness level, policy, tool trace, or any experience text---so it cannot systematically favor a provisioning condition. 

\subsection{Cost Accounting}
Token counts sum the API-reported usage of every executor-side LLM call in a condition---routing calls, self-checks, every escalation attempt, and verification calls for cascades---so policies are charged for their full mechanism; judge tokens are metered separately and never count toward policy cost. Latency is wall-clock per task, summed across attempts, and includes tool and simulator execution. Because absolute costs depend on the serving stack, the main text reports both normalized to Full-$K_5$ within each domain; absolute reference values appear in Appendix~\ref{app:full_results}.

\subsection{Repetition and Randomness}
Every condition runs three independent repetitions; tables report the mean $\pm$ population std of the three run means. Scenarios are deterministic replays, so nondeterminism enters only through model sampling. Execution-derived levels are estimated from the pooled construction-split runs; policies are evaluated on the disjoint test split.

\section{Map Estimation and Statistical Analysis}
\label{app:map_construction}

\subsection{Literature-Derived Map}
We set $A_{\mathrm{lit}}(T,K)=P_{\mathrm{lit}}(K\mid T)$ and $\epsilon_{\mathrm{lit}}=0.05$. Under this tolerance, $\mathcal{G}_{\mathrm{lit}}$ selects the modal harness level for all 12 task classes.

\subsection{Execution-Derived Map}

For each task class $T$, $\mu_{\mathrm{exec}}(T,K)$ is the mean score over the five construction tasks and three runs. The execution-derived map selects the lowest harness level whose score is within $\epsilon_{\mathrm{exec}}=0.05$ of the best observed level. Table~\ref{tab:stability} reports the selected levels and their stability.

\subsection{Tolerance Sensitivity}
\label{app:eps_sweep}
Table~\ref{tab:eps_sweep} sweeps $\epsilon_{\text{exec}}$. Grid floors are invariant for $\epsilon\le0.05$ and Liquid changes only two classes between 0 and 0.05; test accuracy is flat within noise across 0.05--0.15 while token cost varies $<$10\%. The paper's operating point $\epsilon{=}0.05$ sits on this plateau, so no conclusion depends on the tolerance choice.

\begin{table}[t]
\caption{Sensitivity of the execution-derived map to $\epsilon_{\text{exec}}$. ``$\Delta$floors'' counts classes whose selected level differs from $\epsilon{=}0.05$; accuracy/tokens are test-split lookup results.}
\label{tab:eps_sweep}
\centering
\small
\begin{tabular}{lcccccc}
\toprule
& \multicolumn{3}{c}{Liquid} & \multicolumn{3}{c}{Grid} \\
\cmidrule(lr){2-4}\cmidrule(lr){5-7}
$\epsilon$ & $\Delta$fl. & Acc $\pm$ std & Tok & $\Delta$fl. & Acc $\pm$ std & Tok \\
\midrule
0.00 & 2 & $0.659\pm0.021$ & 14{,}738 & 0 & $0.762\pm0.023$ & 11{,}829 \\
0.02 & 2 & $0.659\pm0.021$ & 14{,}738 & 0 & $0.762\pm0.023$ & 11{,}829 \\
0.05 & 0 & $0.670\pm0.020$ & 14{,}009 & 0 & $0.762\pm0.023$ & 11{,}829 \\
0.10 & 3 & $0.673\pm0.022$ & 14{,}246 & 1 & $0.749\pm0.024$ & 12{,}253 \\
0.15 & 4 & $0.671\pm0.024$ & 12{,}851 & 2 & $0.742\pm0.035$ & 13{,}406 \\
\bottomrule
\end{tabular}
\end{table}

\subsection{Execution-derived Level Stability}
Table~\ref{tab:stability} resamples the five construction tasks per class (1{,}000 bootstrap draws) and reports the probability of re-selecting the adopted floor, plus the set of floors reachable by leave-one-task-out. In 19 of 24 class$\times$domain cells the adopted floor is re-selected with probability $\ge0.7$; instability concentrates where adjacent levels are within tolerance of each other (e.g., Liquid I-F-S, where K3/K4/K5 all solve the class), i.e., exactly where the choice is least consequential. No resampling ever moves a floor below K2.

\begin{table}[t]
\caption{Execution-derived level stability at $\epsilon{=}0.05$: bootstrap probability of re-selecting the adopted floor over construction tasks, and leave-one-out floor sets.}
\label{tab:stability}
\centering
\small
\setlength{\tabcolsep}{3.6pt}
\begin{tabular}{lcccccc}
\toprule
& \multicolumn{3}{c}{Liquid} & \multicolumn{3}{c}{Grid} \\
\cmidrule(lr){2-4}\cmidrule(lr){5-7}
Class & floor & $P$(floor) & LOO & floor & $P$(floor) & LOO \\
\midrule
I-N-$y$ & K3 & 0.70 & K3       & K4 & 0.91 & K4 \\
I-N-$x$ & K3 & 0.99 & K3       & K5 & 0.99 & K5 \\
I-N-$m$ & K3 & 0.99 & K3       & K3 & 0.99 & K3 \\
I-F-$y$ & K5 & 0.35 & K3/K4/K5 & K3 & 0.75 & K3/K4 \\
I-F-$x$ & K5 & 0.68 & K4/K5    & K5 & 0.73 & K3/K5 \\
I-F-$m$ & K5 & 1.00 & K5       & K5 & 1.00 & K5 \\
A-N-$y$ & K5 & 0.92 & K5       & K5 & 0.92 & K5 \\
A-N-$x$ & K5 & 0.69 & K3/K5    & K3 & 0.88 & K3 \\
A-N-$m$ & K3 & 0.82 & K3       & K5 & 0.98 & K5 \\
A-F-$y$ & K5 & 1.00 & K5       & K5 & 0.99 & K5 \\
A-F-$x$ & K2 & 0.70 & K2/K3    & K5 & 0.68 & K3/K5 \\
A-F-$m$ & K5 & 0.92 & K5       & K5 & 1.00 & K5 \\
\bottomrule
\end{tabular}
\end{table}

\subsection{Statistical Tests}
Paired bootstrap over the 60 test tasks (per-task scores averaged over three runs; $10^4$ resamples): on Liquid, Exec-Lookup vs.\ Full-$K_5$ gives $\Delta{=}{+}0.018$, 95\% CI $[-0.022,+0.061]$---statistically indistinguishable accuracy at 14\% lower token cost, which is the claimed trade. On Grid, Exec-Lookup is genuinely below Full-$K_5$ ($\Delta{=}{-}0.044$, CI $[-0.087,-0.008]$) and genuinely above Lit-Lookup ($\Delta{=}{+}0.098$, CI $[+0.020,+0.183]$), confirming both that maximal provision stays accuracy-optimal on Grid and that execution evidence is needed to calibrate the literature prior.

\subsection{Oracle Definitions}
The Class Oracle selects, per class, the level with the highest test-split mean; the Task Oracle selects the best level per individual task; both use mean scores across the three runs. Both peek at test outcomes and are therefore non-deployable ceilings; neither has a consistent cost, so they appear as accuracy lines only.

\section{Baseline Implementations}
\label{app:baseline_details}

\subsection{Full-\texorpdfstring{$K_5$}{K5}}
The frozen executor with the complete K5 registry on every task; no routing calls.

\subsection{Instruction--Tool Retrieval}
The official ITR package with default configuration (hybrid dense + BM25 + cross-encoder retrieval). The tool corpus is the 14--16 harness tool specs verbatim; the instruction corpus is the K5 system prompt in fragments, with exemplars from successful construction traces (Grid). Per task, retrieved tools map to the minimal cumulative level containing them (retrieval confidence $<0.7$ falls back to full provision); routing itself costs zero LLM tokens.

\subsection{LLM-Route and LLM+Exp}
One tool-free routing call to the same executor, presented with a generic legend of the five levels and instructed to pick the cheapest sufficient level, terminating in \texttt{LEVEL: Kx}. LLM+Exp additionally shows the router a distilled experience playbook: per-class recipe cards distilled by the executor from construction-split trajectories (multi-run scores, best tool sequences with observation snippets, contrastive failures; the pre-registered floor annotations are withheld), synthesized into a $\le$1{,}200-word global playbook. The playbook informs routing only; execution runs on the raw task. Routing tokens count toward cost.

\subsection{AutoMix}
The official AutoMix implementation mapped from a model cascade to a harness cascade: the K3 episode plays the small model, the K5 episode the large one. Confidence uses the official few-shot self-verification prompt at temperature 1.0 with $k{=}8$ samples; routing uses the official 8-bin POMDP meta-verifier trained on construction-split rows (threshold fallback), with per-domain costs set to measured mean tokens. Verification tokens count toward cost, which is why AutoMix is expensive despite reusing episodes.

\subsection{Blind-ESC}
Identical loop, self-check protocol, one-tier-per-failure climbing, finding-carrying, and grading as Map-ESC; the only difference is the seed ($K_1$ vs.\ the mapped floor). It therefore isolates the value of the map from the value of escalation.

\subsection{Reflexion and ExpeL}
Reflexion runs at fixed K5 with the official reflection prompts and at most three attempts; since benchmark rewards are hidden at run time, the retry trigger is a ground-truth-free LLM self-verdict (PASS/FAIL), and reflections read the real tool trace. ExpeL runs at fixed K5 with the official pipeline: 20 induced rules from construction trajectories plus 6 task-similar few-shot exemplars retrieved by sentence-embedding kNN from 50 successful construction traces.

\section{Full Experimental Results}
\label{app:full_results}

\subsection{Aggregate \texorpdfstring{$K_1$--$K_5$}{K1--K5} Sweep}
Table~\ref{tab:sweep} gives the full-corpus sweep (all 240 tasks $\times$ 3 runs). Aggregate performance is not strictly monotonic, but the aggregate is exactly what the per-class analysis shows to be misleading: In Liquid, $K_4$ performs below $K_3$ in aggregate despite improving performance for several individual task classes. In Grid, $K_5$ is the most accurate level and uses fewer tokens than
$K_3$ and $K_4$, although it is not the least costly level overall.

\begin{table}[t]
\caption{Full $K_1$--$K_5$ sweep, all 120 tasks per domain, 3 runs.}
\label{tab:sweep}
\centering
\small
\begin{tabular}{lcccccc}
\toprule
& \multicolumn{3}{c}{Liquid} & \multicolumn{3}{c}{Grid} \\
\cmidrule(lr){2-4}\cmidrule(lr){5-7}
Level & Acc $\pm$ std & Tok & Lat (s) & Acc $\pm$ std & Tok & Lat (s) \\
\midrule
K1 & $0.253\pm0.002$ & 346      & 3.5  & $0.191\pm0.016$ & 1{,}021  & 4.0 \\
K2 & $0.216\pm0.011$ & 866      & 4.3  & $0.197\pm0.004$ & 2{,}012  & 4.5 \\
K3 & $0.581\pm0.008$ & 11{,}617 & 13.7 & $0.502\pm0.020$ & 14{,}716 & 13.6 \\
K4 & $0.560\pm0.013$ & 16{,}923 & 16.6 & $0.566\pm0.011$ & 18{,}179 & 12.8 \\
K5 & $0.680\pm0.015$ & 17{,}794 & 50.2 & $0.773\pm0.019$ & 13{,}265 & 7.7 \\
\bottomrule
\end{tabular}
\end{table}

\subsection{Per-Class Execution Scores}
Table~\ref{tab:perclass} reports the mean accuracy for all ten tasks in each class under $K_1$--$K_5$. These full-benchmark results complement, rather than reproduce, the construction-split scores shown in Figure~\ref{fig:task_harness_map}.

\begin{table*}[t]
\caption{Per-class mean accuracy $\pm$ std for every harness level (10 tasks/class, 3 runs).}
\label{tab:perclass}
\centering
\small
\setlength{\tabcolsep}{4pt}
\begin{tabular}{lccccc@{\hskip 14pt}ccccc}
\toprule
& \multicolumn{5}{c}{Liquid} & \multicolumn{5}{c}{Grid} \\
\cmidrule(lr){2-6}\cmidrule(lr){7-11}
Class & K1 & K2 & K3 & K4 & K5 & K1 & K2 & K3 & K4 & K5 \\
\midrule
I-N-S & .00$\pm$.00 & .00$\pm$.00 & .90$\pm$.00 & .88$\pm$.06 & .87$\pm$.03 & .03$\pm$.02 & .03$\pm$.02 & .78$\pm$.02 & .90$\pm$.00 & .88$\pm$.02 \\
I-N-X & .02$\pm$.02 & .04$\pm$.03 & .59$\pm$.05 & .53$\pm$.05 & .46$\pm$.10 & .00$\pm$.00 & .00$\pm$.00 & .23$\pm$.06 & .63$\pm$.05 & .90$\pm$.02 \\
I-N-M & .36$\pm$.06 & .29$\pm$.07 & .90$\pm$.00 & .87$\pm$.05 & .85$\pm$.07 & .23$\pm$.02 & .26$\pm$.01 & .93$\pm$.05 & .93$\pm$.05 & .93$\pm$.05 \\
I-F-S & .15$\pm$.00 & .15$\pm$.00 & .70$\pm$.05 & .65$\pm$.07 & .71$\pm$.02 & .10$\pm$.00 & .12$\pm$.02 & .42$\pm$.02 & .45$\pm$.05 & .44$\pm$.10 \\
I-F-X & .20$\pm$.03 & .14$\pm$.02 & .30$\pm$.06 & .33$\pm$.01 & .36$\pm$.06 & .15$\pm$.04 & .10$\pm$.02 & .26$\pm$.06 & .32$\pm$.07 & .36$\pm$.04 \\
I-F-M & .32$\pm$.05 & .31$\pm$.06 & .34$\pm$.04 & .53$\pm$.04 & .78$\pm$.09 & .10$\pm$.04 & .12$\pm$.02 & .30$\pm$.04 & .29$\pm$.06 & .88$\pm$.02 \\
A-N-S & .28$\pm$.06 & .24$\pm$.05 & .55$\pm$.07 & .45$\pm$.04 & .73$\pm$.09 & .30$\pm$.04 & .28$\pm$.05 & .61$\pm$.05 & .61$\pm$.03 & .95$\pm$.00 \\
A-N-X & .53$\pm$.06 & .42$\pm$.05 & .67$\pm$.02 & .60$\pm$.05 & .72$\pm$.07 & .40$\pm$.04 & .39$\pm$.02 & .69$\pm$.05 & .55$\pm$.11 & .65$\pm$.00 \\
A-N-M & .12$\pm$.02 & .25$\pm$.04 & .83$\pm$.06 & .82$\pm$.05 & .85$\pm$.07 & .37$\pm$.02 & .35$\pm$.04 & .75$\pm$.08 & .80$\pm$.00 & .97$\pm$.05 \\
A-F-S & .22$\pm$.06 & .17$\pm$.02 & .27$\pm$.02 & .15$\pm$.00 & .75$\pm$.08 & .26$\pm$.01 & .22$\pm$.02 & .27$\pm$.04 & .33$\pm$.03 & .83$\pm$.02 \\
A-F-X & .23$\pm$.03 & .28$\pm$.04 & .27$\pm$.06 & .19$\pm$.11 & .23$\pm$.02 & .20$\pm$.00 & .17$\pm$.05 & .37$\pm$.13 & .40$\pm$.00 & .49$\pm$.05 \\
A-F-M & .62$\pm$.05 & .30$\pm$.04 & .65$\pm$.04 & .73$\pm$.13 & .83$\pm$.05 & .15$\pm$.04 & .33$\pm$.02 & .40$\pm$.07 & .57$\pm$.02 & 1.00$\pm$.00 \\
\bottomrule
\end{tabular}
\end{table*}
\subsection{Construction- and Test-Split Analysis}
\label{app:split_analysis}

Figure~\ref{fig:task_harness_map} reports construction-split scores, from which the execution-derived map is estimated. All provisioning policies are evaluated on the disjoint test split in Table~\ref{tab:provisioning}. Re-estimating the map on the test split changes the selected level in 10 of the 24 class--domain pairs, primarily in the classes identified as less stable in Table~\ref{tab:stability}. Using the construction-split map results in a mean class-level test regret of $0.036$, consistent with the gap
between Exec-Lookup and the Class Oracle.


\subsection{Absolute Reference Costs}
Normalization anchors for Table~1 (test split, Full-$K_5$): Liquid $16{,}210$ tokens and $47.2$\,s per task; Grid $13{,}409$ tokens and $7.6$\,s per task. Multiplying Table~1 ratios by these anchors recovers absolute costs; e.g., Exec-Lookup averages ${\approx}14.0$k tokens/$39$\,s on Liquid and $11{,}829$ tokens/$8.6$\,s on Grid.

\section{Additional Analyses}
\label{app:additional_analysis}

\subsection{Routing Granularity and Construction Bias}
The class is the right routing unit for this corpus: family-level routing (floors per template family) reaches 0.634/0.704 (Liquid/Grid) and task-level nearest-neighbor routing 0.639/0.690, both below class-level lookup (0.670/0.762). Preferred harnesses also vary \emph{across} families within 7 of 12 Liquid and 5 of 12 Grid classes, so class-level floors are not an artifact of any single template family; they smooth over family-level sampling noise that finer-grained routing overfits.

\subsection{Escalation Diagnostics}
\label{app:escalation_analysis}
Map-ESC escalates rarely: 11.7\% of Liquid and 13.9\% of Grid test executions trigger the $K_5$ retry (mean 1.12/1.14 attempts per task). Seeds follow the map (Liquid: 105/180 executions start at $K_5$, 60 at $K_3$,
and 15 at $K_2$.), and most below-K5 seeds finish where they started (Liquid: 39 of 60 K3-seeded executions; Grid: 20 of 45)---the self-check acts as tail-risk insurance on a calibrated start, not as a router.

\emph{Step-size ablation: is the direct $K_5$ jump too coarse?} We ran the gradual alternative---$K_{i+1}=\min(K_i{+}1,K_5)$, findings carried at every hop---for three repetitions in both domains. It is dominated: accuracy falls to $0.692\pm0.009$ (Liquid) and $0.762\pm0.010$ (Grid) versus $0.715\pm0.014$ and $0.782\pm0.010$ for the direct jump, at equal cost on Liquid and higher cost on Grid ($+5\%$ tokens, $+8\%$ latency). The gradual paths explain why: of the K3-seeded executions that escalated, only 4 of 19 (Liquid) and 1 of 24 (Grid) were satisfied by the intermediate K4 step---the rest climbed on to $K_5$ anyway, having spent an extra attempt at still-insufficient provision and carried a failed intermediate trace into the final context. The insufficiency that survives a correctly-mapped seed is rarely one tier deep, so the intermediate grant buys little; escalating directly to full provision is simultaneously more accurate and more efficient, which is why Map-ESC uses the two-point design.

\subsection{Alternative Experience Representations}
Consistent with the main text, execution experience helps most as a \emph{measurement} (the map) rather than as text: injecting the distilled playbook into execution at Full-$K_5$ helps Grid (0.810 vs.\ 0.806) but not Liquid (0.648 vs.\ 0.652), and routing-only use of the playbook (LLM+Exp) does not beat the free lookup in either domain. ExpeL's trajectory retrieval is the strongest execution-side use of experience on Liquid (0.731) but costs 44\% more tokens than Map-ESC for a comparable gain.

\subsection{Qualitative Error Analysis}
Four recurring failure modes: (i)~\emph{under-provisioning}: below the floor, the agent either abstains or extrapolates from typical ranges (Liquid I-N-S at K1/K2 scores 0.00---the honest failure the taxonomy predicts); (ii)~\emph{over-provisioning backfire}: with simulation available, agents run counterfactuals instead of reading the trend (Liquid I-N-X drops from 0.59 at K3 to 0.46 at K5; Grid A-N-X from 0.69 to 0.65), and sensitivity sweeps burn the iteration budget; (iii)~\emph{executor-limited classes}: Liquid A-F-X never exceeds 0.35 at any level---transient counterfactual deltas exceed what the executor can orchestrate, and no provisioning policy can fix this; (iv)~\emph{self-check errors}: false passes dominate false escalations, which is why Blind-ESC stalls below K5 on classes whose confident-but-wrong answers never trigger the retry.

\section{Artifact Availability and Responsible Deployment}
\label{app:artifact}

\subsection{Released Artifacts}
We will release both 120-task corpora with ground-truth specifications and splits, the frozen scenario telemetry (CSV), the harness manifests and tool schemas for all five levels, the literature corpus IDs with all three per-model annotations and adjudications, the distilled experience playbooks, evaluation and map-estimation code, and per-task execution records (scores, token/latency/simulator meters, tool-call traces) sufficient to reproduce every table and figure without API access.

\subsection{Proprietary Components}

The liquid-cooling twin is currently proprietary. We will release its frozen trajectories, topology metadata, and recorded tool I/O, which are sufficient to reproduce all reported results. Generating new Liquid trajectories currently requires access to the twin; once it is open-sourced, we will release the complete Liquid environment. The Grid environment, based on Grid2Op and pandapower, is fully open-source.

\subsection{Compute and API Usage}
The two $K_1$--$K_5$ sweeps consumed 17.1M (Liquid) and 17.7M (Grid) executor tokens plus $\sim$2.4M judge tokens; policy and baseline conditions add a comparable amount, for $\sim$75M tokens total across all reported experiments. Qwen experiments ran on a single-node vLLM deployment; simulator compute is negligible (sub-second per solve, metered per task).









\end{document}